\documentclass[conference]{IEEEtran}
\usepackage{times}

\usepackage[numbers]{natbib}
\usepackage{multicol}
\usepackage[bookmarks=true]{hyperref}

\usepackage[]{graphicx}
\usepackage{enumerate}
\usepackage{subfigure}
\usepackage{verbatim}
\usepackage{xcolor}
\usepackage{algorithm}
\usepackage{algorithmicx}
\usepackage[noend]{algpseudocode}
\usepackage{tabularx}
\let\labelindent\relax
\usepackage{enumitem}

\usepackage{amssymb}
\usepackage[cmex10]{amsmath}
\usepackage{mathtools}
\usepackage{amsthm}
\newtheorem{definition}{Definition}
\newtheorem{theorem}{Theorem}
\newtheorem{remark}{Remark}
\newtheorem{example}{Example}

\newcommand{\ljmargin}[2]{{\color{magenta}#1}\marginpar{\color{magenta}\raggedright\footnotesize [LJ]:
#2}}

\graphicspath{{./fig/}}

\newcommand{\DFMT}{DFMT$^*$}
\newcommand{\R}{{\mathbb{R}}}
\newcommand{\N}{{\mathbb{N}}}

\newcommand{\xf}{\mathcal{X}_{\text{free}}}
\newcommand{\xg}{\mathcal{X}_{\text{goal}}}
\newcommand{\xobs}{\mathcal{X}_{\text{obs}}}
\newcommand{\xz}{\mbox{$x_0$}}
\newcommand{\xall}{\mathfrak{X}_{\text{free}}}

\DeclareMathOperator*{\argmin}{arg\,min}

\newcommand{\ics}{\textsc{ICS}}

\newcommand{\per}{P}
\newcommand{\dyn}{\mathcal{D}}
\newcommand{\mpk}{MP$_{\text{kn}}$}
\newcommand{\mpu}{MP$_{\text{un}}$}
\newcommand{\act}{\mathcal{A}}
\newcommand{\xfg}{\mathcal{X}^{\text{guess}}_{\text{free}}}

\newcommand{\xfp}{\mathcal{X}^{\per}_{\text{free}}}
\newcommand{\xics}{\mathcal{X}_{\ics}}

\newcommand{\upd}{U\!}
\newcommand{\xop}{\mathcal{X}^{\per}_{\text{obs}}}
\newcommand{\plcy}{Pol$_{\text{\upd}}$}
\newcommand{\icsa}{\widetilde{\ics}}
\newcommand{\xgi}{\mathcal{X}^{\text{inter}}_{\text{goal}}}
\newcommand{\plcya}{Pol$^{\text{approx}}_{\text{\upd}}$}
\newcommand{\e}{\textsc{end}}
\newcommand{\st}{\textsc{start}}
\newcommand{\ra}{\textsc{range}}

\newcommand{\xm}{\mathcal{X}^{\text{map}}_{\text{free}}}
\newcommand{\xallc}{\mathfrak{X}_{\text{obs}}}
\newcommand{\polu}{\sigma^{*}_\upd}
\newcommand{\xfgfwd}{F^{\upd}\!}
\newcommand{\ctgu}{\tilde{c}^*_\upd}
\newcommand{\pola}{\sigma^{\text{aprx}}_\upd}
\newcommand{\ctga}{\tilde{c}^{\text{aprx}}_\upd}

\begin{document}

\title{Safe Motion Planning in Unknown Environments:\\ Optimality Benchmarks and Tractable Policies}


\author{\authorblockN{Lucas Janson}
\authorblockA{Department of Statistics\\
Harvard University\\
Cambridge, Massachusetts 02138\\
Email: ljanson@fas.harvard.edu}
\and
\authorblockN{Tommy Hu}
\authorblockA{Department of Mechanical Engineering\\
Stanford University\\
Stanford, California 94305\\
Email: hutommy@stanford.edu}
\and
\authorblockN{Marco Pavone}
\authorblockA{Department of Aeronautics and Astronautics\\
Stanford University\\
Stanford, California 94305\\
Email: pavone@stanford.edu}}


%

\maketitle

\begin{abstract}
This paper addresses the problem of planning a safe (i.e., collision-free) trajectory from an initial state to a goal region when the obstacle space is a-priori unknown and is incrementally revealed online, e.g., through line-of-sight perception. Despite its ubiquitous nature, this formulation of motion planning has received relatively little theoretical investigation, as opposed to the setup where the environment is assumed known. A fundamental challenge is that, unlike motion planning with known obstacles, it is not even clear what an optimal policy to strive for is. Our contribution is threefold. First, we present a notion of optimality for safe planning in unknown environments in the spirit of comparative (as opposed to competitive) analysis, with the goal of obtaining a benchmark that is, at least conceptually, attainable. Second, by leveraging this theoretical benchmark, we derive a pseudo-optimal class of policies that can seamlessly incorporate any amount of prior or learned  information while still \emph{guaranteeing} the robot never collides. Finally, we demonstrate the practicality of our algorithmic approach in numerical experiments using a range of environment types and dynamics, including a comparison with a state of the art method. A key aspect of our framework is that it automatically and implicitly weighs exploration versus exploitation in a way that is optimal with respect to the information available.
\end{abstract}

\IEEEpeerreviewmaketitle

\section{Introduction}
\label{sec:intro}
This paper addresses the problem of planning a safe (i.e., collision-free)
trajectory from an initial state to a goal region when the obstacles
in between are a priori unknown and are instead revealed
online, e.g., through line-of-sight perception. A fundamental challenge
is that, unlike motion planning with known obstacles or obstacles
whose uncertainty is fully modeled probabilistically, when parts of
the configuration space are simply unknown it is not even clear what
an optimal policy to strive for (through, e.g., asymptotic
convergence) is. One of the main goals of this paper is to address
this shortcoming in the literature by defining a notion of 
optimality for use as a benchmark and also for conceptual
guidance. We then follow this conceptual guidance to propose a novel
algorithm for planning in unknown environments that produces low-cost
solutions by flexibly incorporating side information about the
environment, is guaranteed to be collision-free (even if the side
information is incorrect), and requires on the order of 0.5--1s of (serial) computation time per action.

{\em Related Work}: Most algorithms for motion planning assume full knowledge of the environment, and can not be used, at least directly, to find a motion plan within an incomplete map \cite{LaValle2006,LaValle2011}. A common heuristic approach is to set temporary goals along the way to the goal region, and re-compute in a receding horizon fashion motion plans within the portion of the environment that has been observed (a technique also referred to as re-planning). An instantiation of this approach is represented by frontier-based exploration, in which the temporary goals are placed at the boundary between the observed and unobserved environment (e.g., on the nearest frontier \cite{Yamauchi1997}; see also \cite{BurgardMoorsEtAl2005} for more sophisticated heuristics for temporary goal placements). Such a receding horizon approach is also quite effective in the presence of moving obstacles or, more in general, dynamic environments \cite{HsuKindelEtAl2002,FergusonKalraEtAl2006, BergOvermars06}. 

\begin{figure}[t]
\centering
	\includegraphics[width=0.42\textwidth]{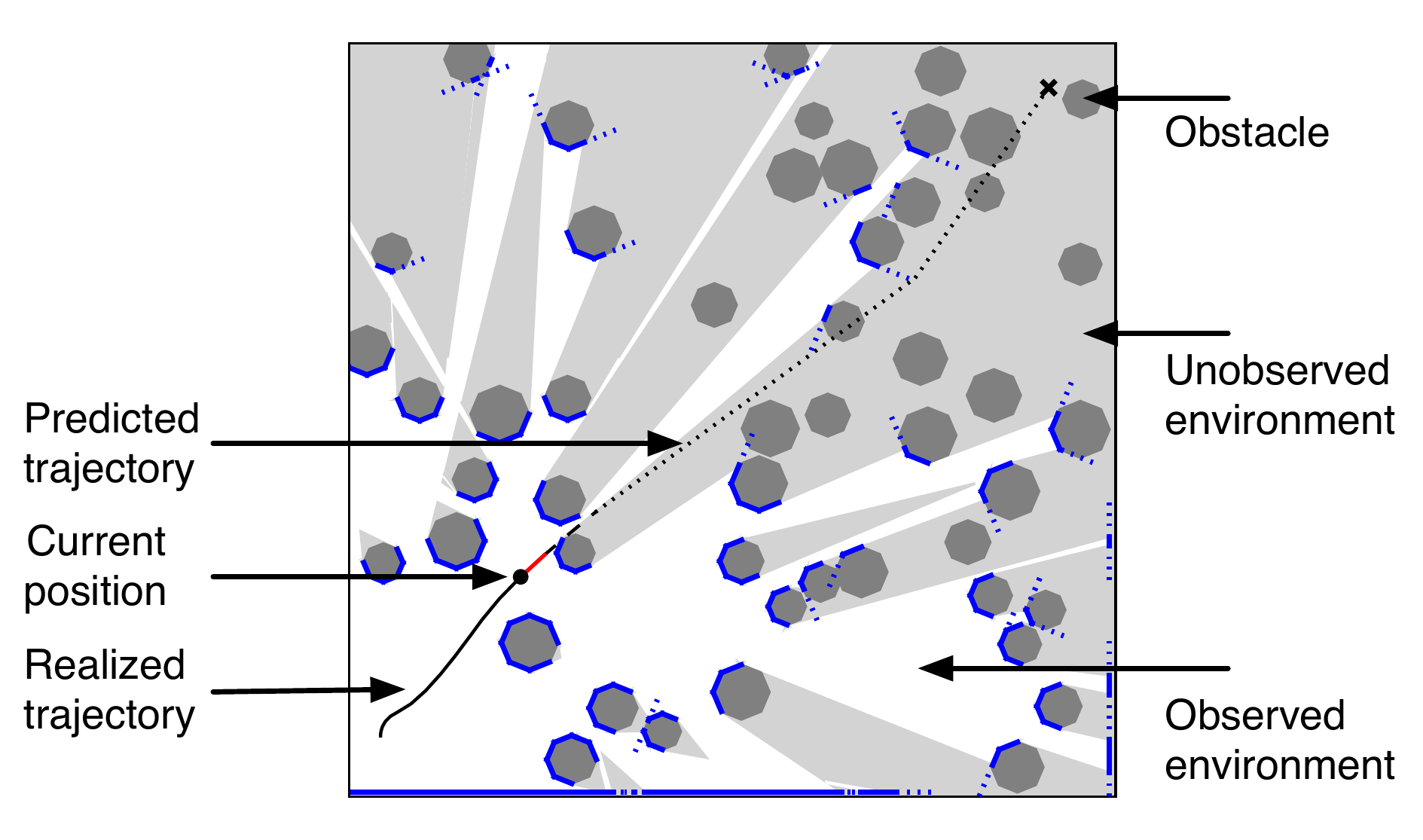}
		\caption{Planning in unknown environments: as the robot moves (in this example, with double-integrator dynamics and line-of-sight perception), it needs to weigh exploration versus exploitation in a way that is best with respect to the information available. In particular, the robot should never take an action that leads into an inevitable collision state with respect to any unobserved part of the environment.}
	\label{fig:intro} 
\end{figure}
If one considers a kinodynamic model of a robot (as is the case in this paper), an additional consideration is how to ensure safety when the environment is only partially known. An important concept in this regard is the notion of inevitable collision states (ICS) \cite{FraichardAsama2004}, states for which, no matter what future control inputs the robot applies, a collision with an obstacle must eventually occur. To ensure safety, the robot should then never  take an action that leads into an inevitable collision state \emph{with respect to any unobserved part of the environment} (see Figure \ref{fig:intro}). The notion of planning in unknown environments under ICS constraints has been studied by a number of authors, see, e.g.,  \cite{FrazzoliDahlehEtAl2002,FraichardAsama2004, PettiFraichard2005, BekrisKavraki2007}, with a focus on making the computation of ICS tractable.

An additional important consideration is how to reason about actions and/or observations that {\em might} be taken in the unobserved environment \cite{HengGotovosEtAl2015}. This is directly related to the notion of selecting actions that best weigh exploration versus exploitation as a function of available information, such as from perception. A possible approach is to bias motion plans toward ``next best view" states \cite{Connolly1985}, which capture the notion of amount of unobserved environment that can be seen at a future state. For example, this idea is infused within a receding horizon planning scheme in \cite{BircherKamelEtAl2016}. A conceptually similar idea is to choose temporary goals that reward exploration \cite{HengGotovosEtAl2015}. More in general, one can frame the problem of planning in unknown environments as a partially observable Markov decision process, where the partial observability is with respect to the environment \cite{RichterVega2018}. To make the  problem tractable, one needs to consider a number of approximations, for example, replacing collision avoidance constraints with penalties on collision probabilities (possibly learned \cite{RichterVega2018}).

\emph{Statement of Contributions:}  Despite the wealth of algorithms available for planning under uncertainty, and the well-established theoretical foundations for planning in known environments (e.g., \cite{KaramanFrazzoli2011,JansonSchmerlingEtAl2015,LiLittlefieldEtAl2016}), the problem of planning in unknown environments has received relatively little theoretical investigation. Accordingly, the contribution of this paper is threefold. First, we present a notion of optimality for planning in unknown environments in the spirit of {\em comparative} \cite{KoutsoupiasPapadimitriou00} (as opposed to competitive \cite{BorodinEl-Yaniv05,FleischerKamphansEtAl2008}) analysis, with the goal of obtaining a benchmark that is, at least conceptually, attainable. Second, by leveraging the theoretical benchmark and combining the aforementioned notions of re-planning, ICS, and forward-looking biasing in a self-contained framework, we derive a pseudo-optimal class of policies that can seamlessly incorporate any amount of prior or learned  information   while still \emph{guaranteeing} the robot never collides. Finally, we demonstrate the practicality of our algorithmic approach in numerical experiments using a range of environment types and robot dynamics, including a comparison with a state of the art method for planning in unknown environments \cite{RichterVega2018}. Importantly, computation times are on the order of 0.5--1s of (serial) computation time per action, making it an algorithm amenable to real-time implementation.
  
\emph{Organization:}  This paper is structured as follows. In Section \ref{sec:method} we provide notation for a number of concepts needed to rigorously state the problem of planning in unknown environments. In Section \ref{sec:optimality} we develop a notion of optimality that can be used as a benchmark. In Section 
\ref{sec:policies} we leverage such a notion of optimality to derive a pseudo-optimal class of policies. In Section \ref{sec:experiments} we present results from numerical experiments supporting our statements. Finally, in Section \ref{sec:conc} we draw some conclusions and discuss directions for future work.

\section{Notation}
\label{sec:method}

\begin{figure}[h]
\centering
\includegraphics[width=0.52\linewidth]{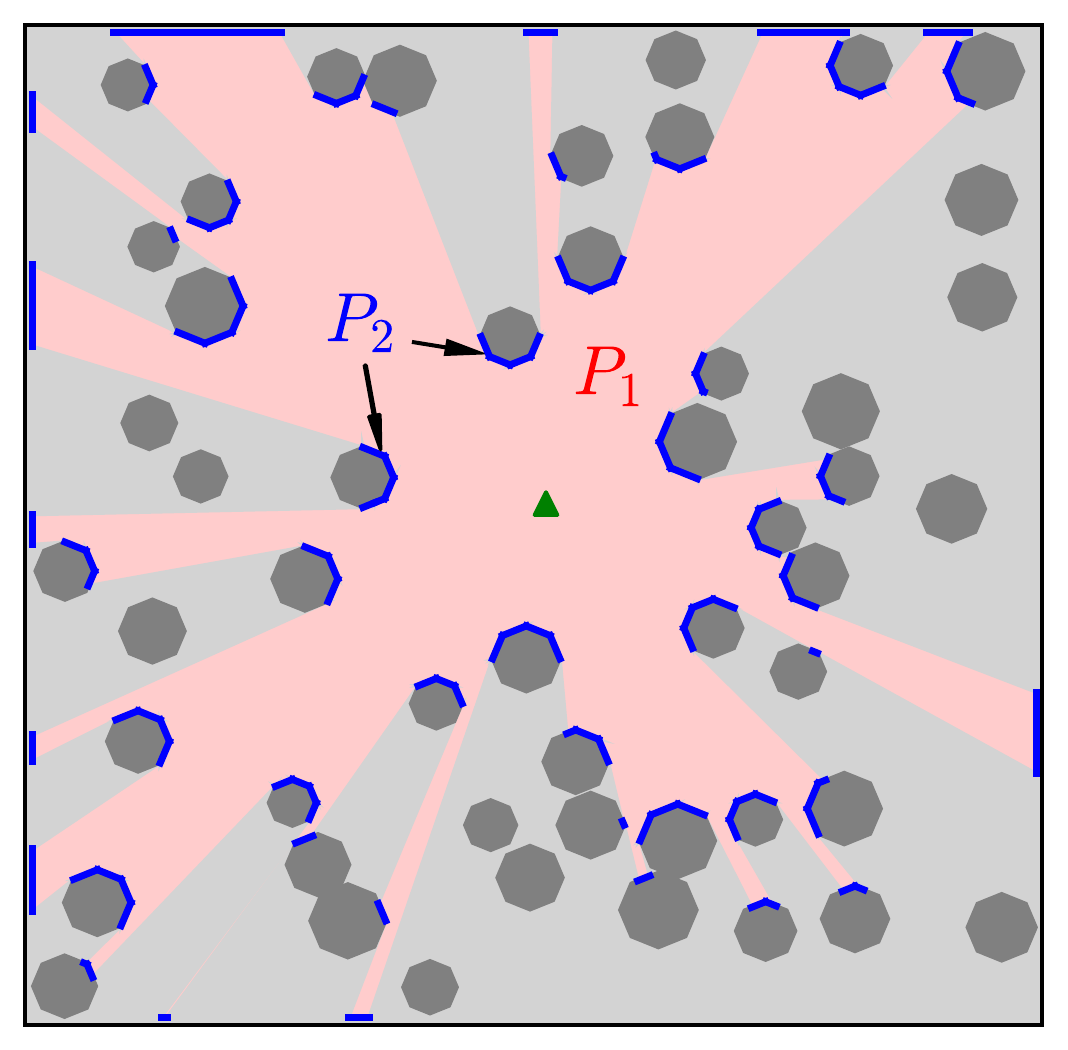}
\caption{\label{fig:p1p2} For a robot (green triangle) with line-of-sight perception, the observed obstacle-free space is given by $\per_1$ in red while the observed obstacle subsets are given by $\per_2$ in blue. Note that $\per_1$ and $\per_2$ are not completely redundant, as the boundary of $\per_1$ excluding $\per_2$ represents unexplored and ostensibly obstacle-free frontier.}
\end{figure}

This paper concerns planning trajectories for robots with dynamics and perception through static configuration spaces with obstacles, and we will need carefully-defined notation for all these concepts. First the robot's configuration space: we assume it lies in $[0,1]^d$, and it is characterized by its obstacles $\xobs\subset [0,1]^d$, or more generally the subset of $[0,1]^d$ which the robot is not allowed to enter. The remainder of the configuration space which the robot is free to traverse is denoted $\xf = [0,1]^d\setminus\xobs$ and we will assume $\xf$ is a closed set. We will sometimes informally refer to the configuration space with obstacles defined equivalently by $\xobs$ or $\xf$ as the environment. In particular, this paper deals with the case of unknown environment, necessitating notation for the set of all possible $\xf$: $\xall = \{\mathcal{X}\subset [0,1]^d : \mathcal{X} \text{ closed}\}$ and all possible $\xobs$: $\xallc = \{\mathcal{X}\subset [0,1]^d : [0,1]^d\setminus\mathcal{X} \text{ closed}\}$. Define the robot's initial state as $\xz\in\xf$ and its goal region as $\xg\subset\xf$, where the set $\xg$ will also be assumed closed. Next, we define the dynamics $\dyn$ as the set of differential equations governing the robot's motion. Let $\sigma:[0,T]\rightarrow[0,1]^d$ denote a finite-time trajectory of duration $T$, with $\Sigma_\dyn$ the set of all dynamically-feasible trajectories under $\dyn$. Let $\st(\sigma):=\sigma(0)$ and $\e(\sigma):=\sigma(T)$, and let $\ra(\sigma)$ denote the curve in $[0,1]^d$ traced out by $\sigma$.
Let $c:\Sigma_\dyn \rightarrow \R_{\ge 0}$ be an additive cost function on the set of finite-time dynamically-feasible trajectories.
Another important component in this paper is the robot's perception mechanism/function $\per:\xall\times[0,1]^d \rightarrow \xall\times \xallc$, which takes in the environment and robot state and $\per_1$ returns a (usually local) set $\xfp$ of collision-free states around the robot's state $x\in[0,1]^d$ and $\per_2$ returns the observed open (except on the boundaries of $[0,1]^d$) subset $\xop$ of $\xobs$ (while many robot sensors will only be able to perceive obstacle boundaries, which are closed, we will assume such sensors instead return a very thin open set along the obstacle boundary).\footnote{$\per(\xf,x)$ may be defined arbitrarily for $x\notin\xf$.}\footnote{$\per$ could also take as its second argument an entire trajectory---everything in this paper would still go through with only minor notational changes.} Note that $\per_2$ is not just redundantly returning $[0,1]^d\setminus\xfp$, as it makes the important distinction between obstacle boundary, which the robot must avoid, and unexplored frontier, which the robot should explore in order to reach its goal; see Figure~\ref{fig:p1p2}. 

The first argument to $\per$ is the configuration space, so the actual perception of the robot at a state $x$ navigating an environment $\xf$ will always be given by $\per(\xf,x)$, but importantly, the explicit dependence on the first argument will also allow us to reason about hypothetical future perception in a postulated environment $\xfg$ which need not correspond to the true environment $\xf$. It will also be useful to have a function $\ics:\xall\rightarrow\xall$, which takes a set of collision-free states $\xf$ and returns the set of \emph{inevitable collision states} (ICS) $\xics$ defined as states $x$ for which there exists $T_x<\infty$ such that a robot in state $x$ and with dynamics $\dyn$ is guaranteed to leave $\xf$ before time $T_x$ (for notational simplicity we suppress the dependence on the dynamics $\dyn$). Throughout this paper we will assume no noise, that is, there is no stochasticity in the dynamics $\dyn$, the perception $\per$, and the robot is assumed to always know its current state $x$. We will also assume that $\xg\cap\ics(\xf) = \emptyset$, although this is not really necessary. Finally, let $[N] := \{1,\dots,N\}$.

\section{A Notion of Optimality for Planning \\in Unknown Environments}\label{sec:optimality}
\subsection{Benchmarking Motion Planning in Unknown Environments}
A significant algorithmic challenge and central contribution of this
paper is how to define the objective for motion planning in unknown
environments. As shorthand and to distinguish the settings of known and unknown environments, we will refer to motion
planning in unknown environments as \mpu{} and motion planning in known
environments as \mpk{}. Pedagogically, we find it useful to compare with the
known-environment setting, where the problem statement is clear:
\begin{quote}
Find a minimum-cost collision-free trajectory from $\xz$ to $\xg$.
\end{quote}
Assuming a collision-free and dynamically-feasible trajectory exists, the minimum cost for the \mpk{} problem with given $\dyn$, $c$, $\xf$, $\xz$, and $\xg$ is given by (where dependence on $\dyn$ and $c$ have been suppressed to streamline the notation)
\begin{equation}
\label{eq:mpkcont}
c^*(\xf,\xz,\xg) = 
\smashoperator{
\min_{\substack{\sigma\in\Sigma_\dyn,\\ \ra(\sigma)\subset\xf,\\ \st(\sigma)=\xz,\;\;\e(\sigma)\in\xg}}}
\; c(\sigma)
\end{equation}
Parsing the constraints: the first line enforces dynamic feasibility,
the second line ensures the trajectory is collision-free, and the
third line requires the trajectory to start at $\xz$ and end in
$\xg$. As achieving $c^*(\xf,\xz,\xg)$ exactly is rarely computationally
possible, actual \mpk{} algorithms usually target something weaker,
such as feasibility (e.g., \cite{LaValleKuffner2001}) or asymptotic
optimality (e.g., \cite{WebbBerg2013, SchmerlingJansonEtAl2015, SchmerlingJansonEtAl2015b, LiLittlefieldEtAl2016}), but it seems
indisputable that $c^*(\xf,\xz,\xg)$ is the gold standard against which any \mpk{} algorithm can be compared. Before getting to \mpu{}, let us rewrite
\eqref{eq:mpkcont} in a way that explicitly acknowledges that in practice most planning problems are dealt with in discrete time. As such, we can use $\dyn$ to
define an action-generating function
$\act_\dyn:[0,1]^d\rightarrow \Sigma_\dyn$ which gives the closed set
of actions (dynamically-feasible finite-length trajectories) the robot
may follow starting from any given state.\footnote{For instance, these actions could
be traces through a prespecified short time or distance of solutions
to the differential equations in $\dyn$ using parameters defined by a
set of low-level controls such as steering angle or acceleration; we
abstract this all away for notational convenience to deal only with
actions.} We can now write the discrete-time analogue of \eqref{eq:mpkcont} as
\begin{equation}
\label{eq:mpkdisc}
c^*(\xf,\xz,\xg) = 
\smashoperator{\min_{\substack{\sigma_1,\dots,\sigma_N \text{
      s.t. } N\in\N,\vspace{0.17cm}\\
    \sigma_k\in\act_\dyn\big(\e(\sigma_{k-1})\big)\;\,\forall k\in[N],\\
    \ra(\sigma_k)\subset\xf\;\,\forall k\in[N],\\
    \e(\sigma_N)\in\xg}}}
\quad\quad\;\;\; \sum_{k=1}^Nc(\sigma_k)
\end{equation}
As written, the minimization is over $N$ as well as the $\sigma_k$, but in practice $N$ will be fixed to some large value (an upper-bound for the number of actions in the optimal trajectory) and $\act_\dyn$ will be augmented to always include a null action which takes zero time, has zero cost, and does not change the robot's state. This allows the robot to take any number of null actions once it reaches $\xg$, so that $c^*$ can accommodate optimal trajectories shorter than $N$ actions. 
Parsing the constraints: the first line divides the trajectory into
a finite number of actions $\sigma_k$, the second line requires
each action to be dynamically feasible (for notational simplicity, define $\e(\sigma_0):=x_0$), the third line requires each
action to be collision-free, and the final line requires the final action to end in $\xg$. 

Despite the work in \mpu{} reviewed earlier, we are not aware of any analogue
to $c^*(\xf,\xz,\xg)$ for \mpu{}, i.e., a gold standard against which to compare an algorithm's performance and measure its suboptimality. 
Equation~\eqref{eq:mpkdisc} is also useful in that if one could find a minimizer, that trajectory would be optimal for discretized \mpk{}. For these reasons we seek an \mpu{} analogue. First note that $c^*(\xf,\xz,\xg)$ will generally represent an unattainable objective for \mpu{}, and thus does not itself represent an ideal benchmark for \mpu{}. 
\begin{example} Consider a robot with a limited sensing radius and bounded speed and acceleration (which can be applied in any direction, that is, unlike a car, the robot has no `forward' direction) so that the stopping distance from its top speed is greater than its sensing radius. Then the only way to achieve $c^*(\xf,\xz,\xg)$ when $\xobs = \emptyset$ is for the robot to immediately accelerate to maximum speed in the direction of the nearest point in $\xg$ and proceed until it reaches $\xg$. However, this trajectory clearly is not collision-free if there is an obstacle along that path, and in fact even if the robot chooses to follow this trajectory until an obstacle is sensed, it is guaranteed to collide if the obstacle is far enough away from $\xz$. We emphasize that this fact makes such a choice inherently unsafe, since even if there is prior information suggesting there are no obstacles, the robot cannot verify this assertion because of its limited sensing and so may still end up colliding with an unexpected obstacle.\end{example}

The preceding paragraph hints at two important points. First, although it was sufficient in \mpk{} to talk only about trajectories, in \mpu{} new information actually comes in during a trajectory that the robot in general can react to, so it is more appropriate to discuss \emph{online  policies} instead, which are functions that take a priori information about the robot and environment and combine it with a history of perception information gathered by the robot up to a given time to produce the robot's next action. In this paper, we want to find policies which are \emph{safe}.
\begin{definition}[Safe Policy]\label{def:safe}
A safe policy is a function that takes in $\act_\dyn$, $c$, $\per$, $\xz$, $\xg$, and a vector of historical perception information of arbitrary length $\left(\per(\xf,\xz),\dots,\per(\xf,x_k)\right)$ and returns an action that is guaranteed not to leave $\xf$, for any value of $\xf$ that agrees with the historical perception information.
\end{definition}
Note that $\xf$ only appears in the inputs to a policy through the perception function $\per$, so the only information about the environment that the policy can use to ensure safety is gathered online by the robot. Also, policies may in general have a random component, in which case we could talk about the \emph{probability} of the policy returning an action that collides with an obstacle, but in this paper we only deem a policy safe if that probability is zero. The goal of this paper is to establish a framework for computing safe policies that produce (collision-free, by definition) trajectories with as low a cost as possible.

Second, the only way to guarantee a policy is safe is to ensure that no action takes the robot into an inevitable collision state \emph{with respect to any unobserved part of the environment}. This is a necessary condition for a policy to be safe as defined in Definition~\ref{def:safe}, since if the robot ever enters an ICS with respect to an unobserved part of the environment, then the robot will (inevitably) collide for some possible values of the environment (e.g., for $\xf = \bigcup_{j=0}^k\per_1(\xf,x_j)$, which matches the observed environment perceived up until the ICS is reached, but for which $\xobs$ contains all the unobserved environment), contradicting the definition. This is suggestive of the following \mpu{} analogue of Equation~\eqref{eq:mpkdisc} (in addition to suppressing dependence on the dynamics $\act_\dyn$ and the cost function $c$, we also now suppress the dependence on $\per$, since all three of these inputs are normally fixed for a given robot):
\begin{equation}
\label{eq:mpudisc}
\tilde{c}^*(\xf,\xz,\xg) = 
\smashoperator{\min_{\substack{\sigma_1,\dots,\sigma_N \text{
      s.t. } N\in\N,\vspace{0.17cm}\\
    \sigma_k\in\act_\dyn\big(\e(\sigma_{k-1})\big)\;\,\forall k\in[N],\\
    \ra(\sigma_k)\,\bigcap\,\ics\left(\bigcup_{j=0}^{k-1}\per_1\left(\xf,\,x_j\right)\right)=\emptyset\;\,\forall k\in[N],\\
    \e(\sigma_N)\in\xg}}}
\quad\quad\;\;\; \sum_{k=1}^Nc(\sigma_k)
\end{equation}
where the second-to-last constraint line is the only part that changed from
Equation~\eqref{eq:mpkdisc} (we abuse notation slightly by letting $\ics$ now and for the remainder of the paper denote its discrete analogue, defined as the set of states $x$ for which there exists $T_x < \infty$ such that a robot in state $x$ and following any actions in $\act_\dyn$ is guaranteed to leave $\xf$ before time $T_x$). That line states that the $k$th action cannot be in an
ICS with respect to everything that the robot will perceive up through the end of $(k-1)$th action $\sigma_{k-1}$. Note that all the states in $\xobs$ are ICS, so this is at least as strong
as just requiring all the actions to be collision-free, and thus
$\tilde{c}^*\ge c^*$ when the two take
the same arguments. Nevertheless, because any trajectory returned by a safe policy can never enter an ICS (and thus can never violate the second constraint line in Equation~\eqref{eq:mpudisc}), $\tilde{c}^*(\xf,\xz,\xg)$ lower-bounds the cost of any trajectory returned by a safe policy applied in $\xf$. The natural next question is whether Equation~\eqref{eq:mpudisc} is tight---that is, does there exist a safe policy which attains it? 

The answer is a qualified ``yes." As we show next, there is a safe policy that achieves the cost $\tilde{c}^*(\xf,\xz,\xg)$, but that policy depends on $\xf$. A reasonable criticism would be: if we allow the policy to depend on $\xf$, why not have a policy that just returns a minimizer of Equation~\eqref{eq:mpkdisc}, which would attain the even lower cost $c^*$? The answer is that such a policy is not safe as we have defined it, in that it does not in general produce a collision-free trajectory, while the \emph{safety} of the $\xf$-dependent policy we describe next actually does not depend on $\xf$. That is, the policy we are about to define produces a trajectory that is collision-free no matter what, even if $\xf$ is not the true environment. As underscored in Example 1, this distinction is not only conceptually important, but will be practically important when we turn to developing algorithms for real problems, where the true $\xf$ is never known exactly but may be guessed at based on prior knowledge or information gained along the trajectory, making it crucial that the safety of a policy does not depend on perfect knowledge of $\xf$.

Consider a policy that starts with some guess $\xfg$ for $\xf$ and takes actions according to a minimizer of Equation~\eqref{eq:mpudisc}---except with $\xfg$ replacing $\xf$---until it reaches $\xg$ or it first perceives an obstacle boundary not in $\xfg$ (here is where we first see the utility of $\per_2$, since only $\per_1$ was used in Equation~\eqref{eq:mpudisc} but it does not distinguish between observed obstacles and the simple boundary of observed free space). In the latter case---denote such a waypoint by $x_k$---the robot updates $\xfg$ to some $\mathcal{X}^{\text{guess(2)}}_{\text{free}}$ that matches its perception (e.g., by taking the union of the obstacles in $\xfg$ and any obstacle boundary perceived thus far) and then follows a minimizer of Equation~\eqref{eq:mpudisc}---with $\mathcal{X}^{\text{guess(2)}}_{\text{free}}$ now replacing $\xf$, with $x_k$ as its starting point, and still with all perception information up to time $k$ incorporated into the second constraint---until it reaches $\xg$ or it first perceives an obstacle boundary not in $\mathcal{X}^{\text{guess(2)}}_{\text{free}}$, and continues this loop indefinitely. This is a safe policy because by definition the trajectory it produces never enters a state that is ICS with respect to the unobserved part of the environment. Furthermore, in the lucky case when $\xfg = \xf$, an obstacle boundary not in $\xfg$ will never be observed, and the trajectory will simply be a minimizer of Equation~\eqref{eq:mpudisc} and thus, will have cost exactly equal to $\tilde{c}^*(\xf,\xz,\xg)$. The following theorem summarizes what we have shown in this section.

\begin{theorem}\label{thm:mpucost}
For a robot constrained to take actions defined by $\act_\dyn$ in an environment $\xf$ which is unknown to it except through a perception mechanism $\per$, the optimal cost $c$ of any trajectory from $\xz$ to $\xg$ returned by a safe policy is exactly $\tilde{c}^*(\xf,\xz,\xg)$, defined by Equation~\eqref{eq:mpudisc}.
\end{theorem}

\begin{remark}
We stop here to briefly contrast our benchmark with the usual approach of competitive analysis \cite{BorodinEl-Yaniv05}, which would benchmark an online algorithm to its best possible offline performance---in this paper that would mean using $c^*$ as a benchmark instead of $\tilde{c}^*$. But since we have shown that $c^*$ is in general unachievable in \mpu{}, increasing $c^*$ to $\tilde{c}^*$ by imposing constraints that are necessary conditions for any safe \mpu{} trajectory provides a more realistic benchmark. Our approach is similar to that of \emph{comparative} analysis \cite{KoutsoupiasPapadimitriou00}, with the addition that Theorem~\ref{thm:mpucost} also proves that our benchmark is tight.
\end{remark}

\subsection{A Flexible Class of Pseudo-Optimal Policies}
The proof of Theorem~\ref{thm:mpucost}'s tightness in the previous subsection exhibited a class of safe policies that achieve cost exactly equal to $\tilde{c}^*(\xf,\xz,\xg)$ when fed a guess $\xfg$ of $\xf$ that happens to be correct (and because they are safe policies, even if the guess is incorrect they are still guaranteed to never produce a colliding trajectory). This is certainly a desirable property for a safe \mpu{} policy, but perhaps more important from a practical standpoint is ensuring the cost does not depend too dearly on the accuracy of $\xfg$. In simpler terms, we would like our safe policy to achieve \emph{nearly} optimal cost when given a \emph{nearly} correct guess of $\xf$, and to still perform reasonably well even if no prior knowledge about $\xf$ is available. In this subsection we will introduce a flexible class of safe policies that are all pseudo-optimal in that they achieve $\tilde{c}^*(\xf,\xz,\xg)$ when fed a correct guess $\xfg=\xf$, but also allows for more clever methods of updating $\xfg$ as new obstacles are observed online, so that the cost degrades gracefully as $\xfg$ diverges from $\xf$. We still ignore computational constraints in this subsection, but address the approximations needed for computability in the following subsection.

As just alluded to, we first need notation to describe a $\xfg$-updating scheme. Let $\upd:\xall\times \xallc\rightarrow\xall$ take as arguments the thus-far perceived collision-free space $\xfp$ and the thus-far perceived subset of $\xobs$, $\xop$, and return a guess $\xfg$ of what the true $\xf$ might be. A minimal requirement for a reasonable guessing function $U$ would be that its output guess at least agrees with its inputs of historical perception, i.e., $\xfp\subset\upd(\xfp,\xop)$ and $\xop\subset[0,1]^d\setminus\upd(\xfp,\xop)$. This leaves many reasonable choices for $\upd$, such as if there is an outdated map of the environment $\xm$, then a reasonable choice might be $\upd(\xfp, \xop) = \left(\xm\cup\xfp\right) \setminus \xop$, which would produce highly accurate guesses if $\xm$ is pretty close to $\xf$, while still being able to adapt to unexpected obstacles. At the other end of the spectrum, if very little is known about the environment a priori, since we are only considering safe policies anyway, it may make sense to take an aggressive strategy and simply set $\upd(\xfp,\xop) = [0,1]^d\setminus\xop$. Our generic formalism for $\upd$ allows for both extremes and everything in between.

With $\upd$ in hand, the problem simplifies in the sense that there is a well-defined ``best" policy which assumes its guess at the environment is correct, still constrains itself to never enter an ICS with respect to unknown parts of the environment, and is also forward-looking in that it takes into account the consequences of its current action on all future actions. From a given state $x$ with $\xfp$ the union of $\per_1$'s thus far and $\xop$ the union of $\per_2$'s so far, define the optimal cost-to-go when using a given updating function $\upd$ as
\begin{equation}\label{eq:costtogo}
\ctgu\left(x,\left(\xfp,\xop\right)\!,\xg\right) \,\; = \,\;
\smashoperator{\min_{\substack{\sigma_1,\dots,\sigma_N \text{
      s.t. } N\in\N,\vspace{0.17cm}\\
    \sigma_k\in\act_\dyn\big(\e(\sigma_{k-1})\big)\;\,\forall k\in[N],\\
    \ra(\sigma_k)\,\bigcap\,\ics\left(\xfgfwd_1\left(\xfp,\xop,(\sigma_1,\dots,\sigma_{k-1})\right)\right)=\emptyset\;\,\forall k\in[N],\qquad\qquad\\
    \e(\sigma_N)\in\xg}}}
\quad\quad\;\;\; \sum_{k=1}^Nc(\sigma_k)
\end{equation}
where again for notational simplicity define $\e(\sigma_0):=x$, and the function $\xfgfwd:\xall\times\xallc\times\{(\sigma_1,\dots,\sigma_k):\sigma_j\in\Sigma_\dyn\;\forall j\in[k],\,k\in\N\}\rightarrow\xall\times\xallc$ takes in the current state of perception in $\xfp$ and $\xop$, as well as a sequence of future actions $\{\sigma_1,\dots,\sigma_k\}$, and pushes forward the perception function $\per$ and the updating function $\upd$ along the hypothetical trajectory $\sigma_1,\dots,\sigma_k$, returning what the robot would have perceived by the end of $\sigma_k$ assuming $\upd$ produces a correct guess of the environment. Function $\xfgfwd$ can be explicitly defined recursively as:

\begin{align}\label{eq:xfgfwd}
\xfgfwd\left(\xfp,\xop,\emptyset\right) &= \left(\xfp,\xop\right),\\
\xfgfwd_1\left(\xfp,\xop,(\sigma_1,\dots,\sigma_k)\right) &=\nonumber\\
&\hspace{-3.4cm}\xfgfwd_1\left(\xfp,\xop,(\sigma_1,\dots,\sigma_{k-1})\right)\,\bigcup\nonumber\\
&\hspace{-3cm}\per_1\left(\upd\left(\xfgfwd\left(\xfp,\xop,(\sigma_1,\dots,\sigma_{k-1})\right)\right),\e(\sigma_k)\right),\nonumber\\
\xfgfwd_2\left(\xfp,\xop,(\sigma_1,\dots,\sigma_k)\right) &=\nonumber\\
&\hspace{-3.4cm}\xfgfwd_2\left(\xfp,\xop,(\sigma_1,\dots,\sigma_{k-1})\right)\,\bigcup\nonumber\\
&\hspace{-3cm}\per_2\left(\upd\left(\xfgfwd\left(\xfp,\xop,(\sigma_1,\dots,\sigma_{k-1})\right)\right),\e(\sigma_k)\right).\nonumber
\end{align}
Using Bellman's equation and the recursive nature of $\xfgfwd$, $\ctgu$ can also be written recursively as
\begin{align}\label{eq:costtogobell}
\ctgu\!\left(x,\left(\xfp,\xop\right)\!,\xg\right) &= \\
&\hspace{-2.9cm}\smashoperator{\min_{\substack{\sigma\in\act_\dyn(x),\\
    \hspace{1cm}\ra(\sigma)\,\bigcap\,\ics\left(\xfp\right)=\emptyset}}}
\hspace{0.3cm} c(\sigma) + \ctgu\!\left(\e(\sigma),\xfgfwd\left(\xfp,\xop,\{\sigma\}\right)\!,\xg\right)\nonumber
\end{align}
and we can in turn explicitly write the optimal policy similarly
\begin{align}\label{eq:polu}
\polu\left(x,\left(\xfp,\xop\right),\xg\right) &= \\
&\hspace{-3.2cm}\smashoperator{\argmin_{\substack{\sigma\in\act_\dyn(x),\\
    \hspace{1.2cm}\ra(\sigma)\,\bigcap\,\ics\left(\xfp\right)=\emptyset}}}
\hspace{0.1cm} c(\sigma) + \ctgu\!\left(\e(\sigma),\xfgfwd\left(\xfp,\xop,\{\sigma\}\right)\!,\xg\right)\!.\nonumber
\end{align}
($\sigma$ as an optimization variable here represents a single action, in contrast to Equation~\eqref{eq:mpkcont} where it is a full trajectory ending at $\xg$.) Note that as long as $\upd$ is chosen such that it returns the true $\xf$ until it is given contradicting perception information, the cost of following $\polu$ from $\xz$, namely, $\ctgu\left(\xz,\per\left(\xf,\xz\right)\!,\xg\right)$, will exactly equal the optimal \mpu{} cost proved in Theorem~\ref{thm:mpucost}: $\tilde{c}^*(\xf,\xz,\xg)$. This can be seen by noting that if $\upd$ always returns $\xf$, then the second-to-last constraint of Equation~\eqref{eq:costtogo} matches that of Equation~\eqref{eq:mpudisc}, and everything else matches as well. 

We do not try to characterize the gap between $\ctgu$ and $\tilde{c}^*$ for general $\upd$, as such a theoretical investigation would be beyond the scope of this paper, and a simulation experiment would be computationally prohibitive due to the cost of computing $\tilde{c}^*$. It is still useful, however, to have it as a benchmark since it helps in the conceptual development of our algorithm, and it at least provides the pseudo-optimality property of the policy class $\polu$.

\section{An Approximately Pseudo-Optimal\\ Class of Policies}\label{sec:policies}
\subsection{Generic Approximations}
We now turn to the implementation and computational tractability of employing the policy $\polu$. Despite the fairly compact recursive representation of $\ctgu$ in Equation~\eqref{eq:costtogobell}, performing dynamic programming would be prohibitively expensive due to the exponential (in $N$) size of the search space. But luckily, there are a few approximations/simplifications we can make that drastically improve the computational properties of $\polu$, and we have found in practice that our approximate $\polu$ still performs quite well (as discussed in Section \ref{sec:experiments}). We list all these approximations next, which together provide  an \emph{approximately} pseudo-optimal class of policies that are also computationally tractable. In the next subsection, we will give the implementation details for these approximations that we found to work well in our numerical experiments.

\vspace{0.1truecm}
\noindent {\em Approximation \#1}: First, $\ics$ is outer-bounded by $\icsa$ which checks only a few stopping actions (e.g., maximum deceleration with no turning angle, maximal left turning angle, and maximal right turning angle). Note that because this approximation is a tight outer-bound, it does not change the guarantee that our policy never results in collision.

\vspace{0.1truecm}
\noindent {\em Approximation \#2}: One thing to note about Equation~\eqref{eq:xfgfwd} is that if $\upd$ has the reasonable property that its output $\xfg$ does not change with new perception information as long as that information agrees with $\xfg$, then $\xfgfwd$ simplifies and can be written nonrecursively as
\begin{align}\label{eq:simplexfgfwd}
\xfgfwd_1\left(\xfp,\xop,(\sigma_1,\dots,\sigma_k)\right) &=\\
&\hspace{-1.5cm} \xfp\bigcup_{j=1}^k\per_1\left(\upd(\xfp,\xop),\e(\sigma_j)\right)\nonumber\\
\xfgfwd_2\left(\xfp,\xop,(\sigma_1,\dots,\sigma_k)\right) &=\nonumber\\
&\hspace{-1.5cm} \xop\bigcup_{j=1}^k\per_2\left(\upd(\xfp,\xop),\e(\sigma_j)\right).\nonumber
\end{align}
The reason is that no matter how far into the future the guess is being propagated, since the hypothetical future actions are all assumed to gather perception information from an environment that matches $\upd(\xfp,\xop)$, and new perception information that matches the existing guess does not change the guess, then that guess is the same as just the guess with the current information. This property of $\upd$ is not necessary, but both of the examples given when $\upd$ was introduced satisfy it, and even for some $\upd$ functions that do not, we expect that solving Equation~\eqref{eq:costtogo} with the simplified constraint given by Equation~\eqref{eq:simplexfgfwd} will give a good approximation to solving it with the full constraint for most reasonable choices of $\upd$. Thus we plug in the union in Equation~\eqref{eq:simplexfgfwd} for $\xfgfwd_1$ in the second-to-last constraint line of Equation~\eqref{eq:costtogo}.

\vspace{0.1truecm}
\noindent {\em Approximation \#3}: Having just rewritten the second-to-last constraint line of Equation~\eqref{eq:costtogo} using Equation~\eqref{eq:simplexfgfwd}, we now proceed to simplify it further by reducing the union in Equation~\eqref{eq:simplexfgfwd} to only $\xfp$ and the last perception output, i.e., $\xfp\cup\per_1\left(\upd(\xfp,\xop),\e(\sigma_k)\right)$ if the third argument to $\xfgfwd_1$ has at least one action, or simply $\xfp$ if it has none. That is, for checking ICS for hypothetical future actions, we always keep track of all the \emph{actual} perception information gathered by the robot, but only keep track of the most recent \emph{hypothetical} perception information. We expect this approximation to be quite good in most cases, since the robot will usually be moving forward into unknown space, not backward into already-(hypothetically-)explored space. This approximation gives the optimization problem optimal substructure, which makes it far easier to solve.

\vspace{0.1truecm}
\noindent {\em Approximation \#4}: Until $\xg$ first intersects $\xfp$, we greedily set an intermediate goal region just outside $\xfp$ and only optimize our hypothetical trajectory to reach that instead of $\xg$. Said intermediate goal $\xgi\left(\xfp,\xop,\xg\right)$ is chosen as the collision-free region of the boundary of the unknown environment that minimizes the sum of an approximate cost-to-come from $x$ to $\xgi$ and an approximate cost-to-go from $\xgi$ to $\xg$ through the current guess of the environment. This approximation drastically reduces the search space, especially early in the trajectory, and for simple cost-to-come/cost-to-go approximations adds little computational cost. The intuition behind this approximation is that the robot, without knowing what lies beyond the next obstacle, is unlikely to be able to make good high-level choices and must instead be greedy at low resolution. For instance, if a robot traversing a hallway encounters a fork, with limited information about the environment beyond its immediate vicinity, it might as well choose the hallway which points more towards $\xg$. This approximation does nothing to hinder the main benefit of the forward-looking nature of $\polu$, which is highly-intelligent \emph{local} behavior, like how best to approach a blind corner.
\vspace{0.1truecm}

Combining these approximations gives us an expression for an approximation to $\polu$ that is more computationally tractable:
\begin{align}\label{eq:pola}
\pola\left(x,\left(\xfp,\xop\right),\xg\right) &= \\
&\hspace{-3.6cm}\smashoperator{\argmin_{\substack{\sigma\in\act_\dyn(x),\\
    \hspace{1.2cm}\ra(\sigma)\,\bigcap\,\icsa\left(\xfp\right)=\emptyset}}}
\hspace{0.1cm} c(\sigma) + \ctga\!\left(\e(\sigma),\left(\xfp,\xop,\sigma\right)\!,\xg\right)\!,\nonumber
\end{align}
where the new approximate cost-to-go function $\ctga$ is given by (note it has different inputs from $\ctgu$)
\begin{equation}\label{eq:costtogoapprox}
\ctga\left(x,\left(\xfp,\xop,\sigma_0\right)\!,\xg\right) \,\; = \,\;
\smashoperator{\min_{\substack{\sigma_1,\dots,\sigma_N,\vspace{0.17cm}\\
    \sigma_k\in\act_\dyn\big(\e(\sigma_{k-1})\big)\;\,\forall k\in[N],\\
    \ra(\sigma_k)\,\bigcap\,\icsa\left(\xfp\cup\per_1\left(\upd(\xfp,\xop),\e(\sigma_k)\right)\right)=\emptyset\;\,\forall k\in[N],\qquad\qquad\\
    \e(\sigma_N)\in\xgi\left(\xfp,\xop,\xg\right)}}}
\quad \sum_{k=1}^Nc(\sigma_k).
\end{equation}

\subsection{Implementation Details}
The way we implemented $\pola$ made various choices about the approximations discussed in the previous subsection, which can be categorized into the choice of $\upd$, $\icsa$, $\xgi\left(\xfp,\xop,\xg\right)$, and how we actually computed a solution for Equation~\eqref{eq:pola}. These choices are purely heuristic, and were found through experience and experimentation to work well for a range of problems, but there may be many other good choices as well. In all of our experiments the cost function $c$ was simply the elapsed time of the trajectory.

\vspace{0.1cm}
\noindent {\em Choice of $\upd$}: We chose $\upd$ so that all unobserved space is guessed to be collision-free, except any obstacle boundaries that are already observed but end at unknown territory are extended by a small amount into the unknown. This choice of $\upd$ is optimistic while assuming roughly that ``walls are locally smooth'', and we expect this to be a reasonable choice in almost any setting with little a-priori knowledge of the unknown environment. For observed obstacle boundaries that ended at unknown territory, five evenly-spaced points were marked up to 0.125m from the unknown boundary, and then those points were fit with either a linear or quadratic curve and extended into the unknown until intersection with a known region, up to a maximum length of 0.125m.
  
\vspace{0.1cm}
\noindent {\em Choice of $\icsa$}: Three stopping actions were used: maximal straight-line deceleration in current direction of travel, maximal deceleration while turning left, and maximal deceleration while turning right. 

\vspace{0.1cm}
\noindent {\em Choice of $\xgi$}: Intermediate goals were selected by dividing the unknown frontier into lengths of 0.3m, and the location defined by the midpoint of each frontier segment is used to define the center of a circular candidate $\xgi$ of radius 0.3m. The cost-to-come for each candidate was computed by dynamically-constrained (but not ICS-constrained) FMT$^*$ \cite{SchmerlingJansonEtAl2015,SchmerlingJansonEtAl2015b} and the cost-to-go computed by FMT$^*$ \cite{JansonSchmerlingEtAl2015}, both with sample density of 150 samples/m$^2$ and connection radius of 0.75m. Since FMT$^*$ returns a geometric path and not a kinodynamic trajectory, the FMT$^*$ solution path distance was converted to time by assuming that from the final speed when the robot reaches $\xgi$, the robot maximally accelerates to and continues at maximum speed to $\xg$ along the FMT$^*$ path. 

\vspace{0.1cm}
\noindent {\em Choice of solver for Equation~\eqref{eq:pola}}: Finally, the actual optimization problem was approximated again by dynamically-constrained FMT$^*$ but with the addition of the ICS constraint in the third constraint line of Equation~\eqref{eq:costtogoapprox}. In fact, even that constraint was again simplified by replacing $\ra(\sigma_k)$ with $\e(\sigma_k)$. Because of this simplification and approximations \#2 and \#3 in the previous subsection, the ICS constraint can actually be checked pointwise, making it extremely amenable to sampling-based motion planning algorithms such as FMT$^*$, since each sample can be checked for ICS immediately after being sampled and before any edges are drawn. Note that sampling-based algorithms take a slightly different approach to approximate the solution to Equation~\eqref{eq:pola}, but the sample density and edges between samples implicitly define the discreteness parameters $N$ and the set of actions $\act_\dyn$.

\section{Numerical Experiments}\label{sec:experiments}
In this section we present simulation results on maze, forest, and office building environments to show the versatility of our algorithm in environments of varying complexity. We implemented two dynamics models for experiments. The first is a nonlinear vehicle model with maximum acceleration/braking of 1m/s$^2$, maximum speed of 9 m/s, maximum curvature of 0.13 m, and maximum curvature rate of 7.5 (ms)$^{-1}$. The second is a 2D double integrator dynamics model with maximum acceleration/braking of 1 m/s$^2$ and maximum speed of 6 m/s. Experiments were written in the Julia programming language \cite{BezansonKarpinskiEtAl2012} and run on a Intel Core i5-4300U CPU@1.90GHz quadcore laptop, although the four cores were not used in parallel.

\begin{figure}[h]
\begin{center}
\includegraphics[width=0.95\linewidth]{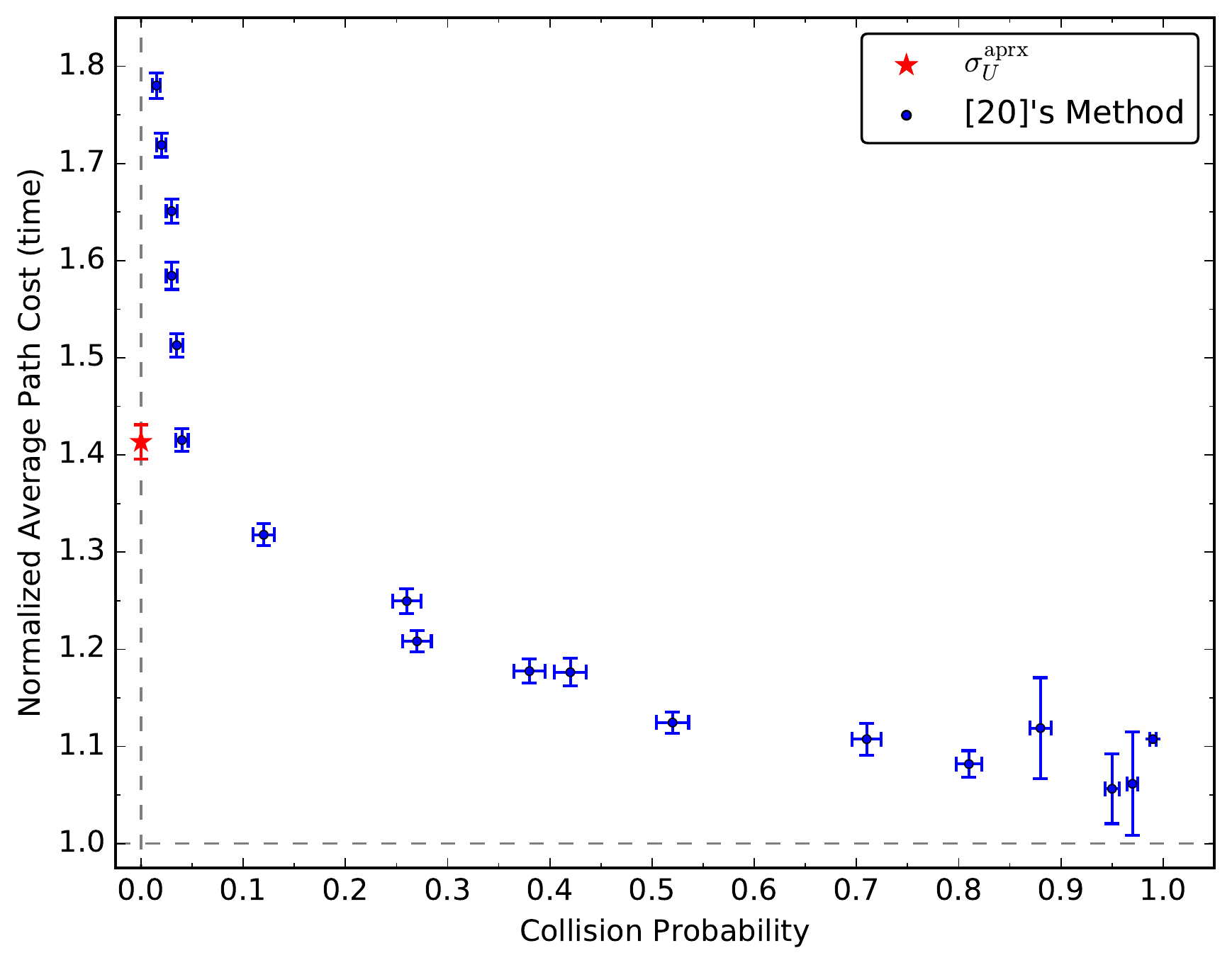}
\end{center}
\caption{\label{fig:pathcost_v_cp} Normalized average path cost versus collision probability of simulations from 500 random hallway maps using our algorithm (red). Path costs are normalized by the cost of the path planned by the kFMT* algorithm with comparable sample density and with full knowledge of the map, and then averaged over the trials. The data points for the \cite{RichterVega2018} method (blue) were produced by running their algorithm on 1000 randomly generated hallway maps with different cost-of-collision values and averaging over the paths that did not result in collision. Their algorithm was trained on 50,000 sample hallway maps and features used were the same as the ones in the original paper, with appropriate hyperparameter tuning. Error bars denote $\pm$1 standard deviation.}
\end{figure}

We first demonstrate the strength of our algorithm in a large-scale comparison with the method of \cite{RichterVega2018}. We ran our algorithm on randomly-generated hallway (maze with no dead ends) maps with the nonlinear vehicle dynamics model and considered the resulting tradeoff between path cost and collision probability. Hallway maps of width 1.2 m were generated by sampling from a Markov chain hallway generator with turn frequency of 0.4. The path cost for connections considered is total time to goal, and the resulting total path cost on each map was normalized by the path cost of the path planned by the dynamically-constrained (but not ICS-constrained) FMT$^*$ algorithm with comparable sample density on the fully known map. Figure \ref{fig:pathcost_v_cp} shows the resulting comparison between our algorithm and that of \cite{RichterVega2018}. By design, our algorithm is guaranteed not to collide and thus only has one data point denoting zero probability of collision. The data point for our algorithm lies below the curve defining the collision probability-path cost tradeoff of \cite{RichterVega2018}'s algorithm, and for their approach to achieve the same path cost as $\pola$ would require roughly a 5\% probability of collision. Additionally, as \cite{RichterVega2018}'s method's probability of collision approaches zero, the cost begins to climb rapidly, far above that of $\pola$. This outperformance can most readily be explained by the forward-looking nature of $\pola$, as it optimizes with respect to future actions and perception, in contrast to the method of \cite{RichterVega2018}.

\begin{figure}[t!]
\centering     
\subfigure[]{
\centering
\label{fig:maze_b}\includegraphics[width=0.3\linewidth]{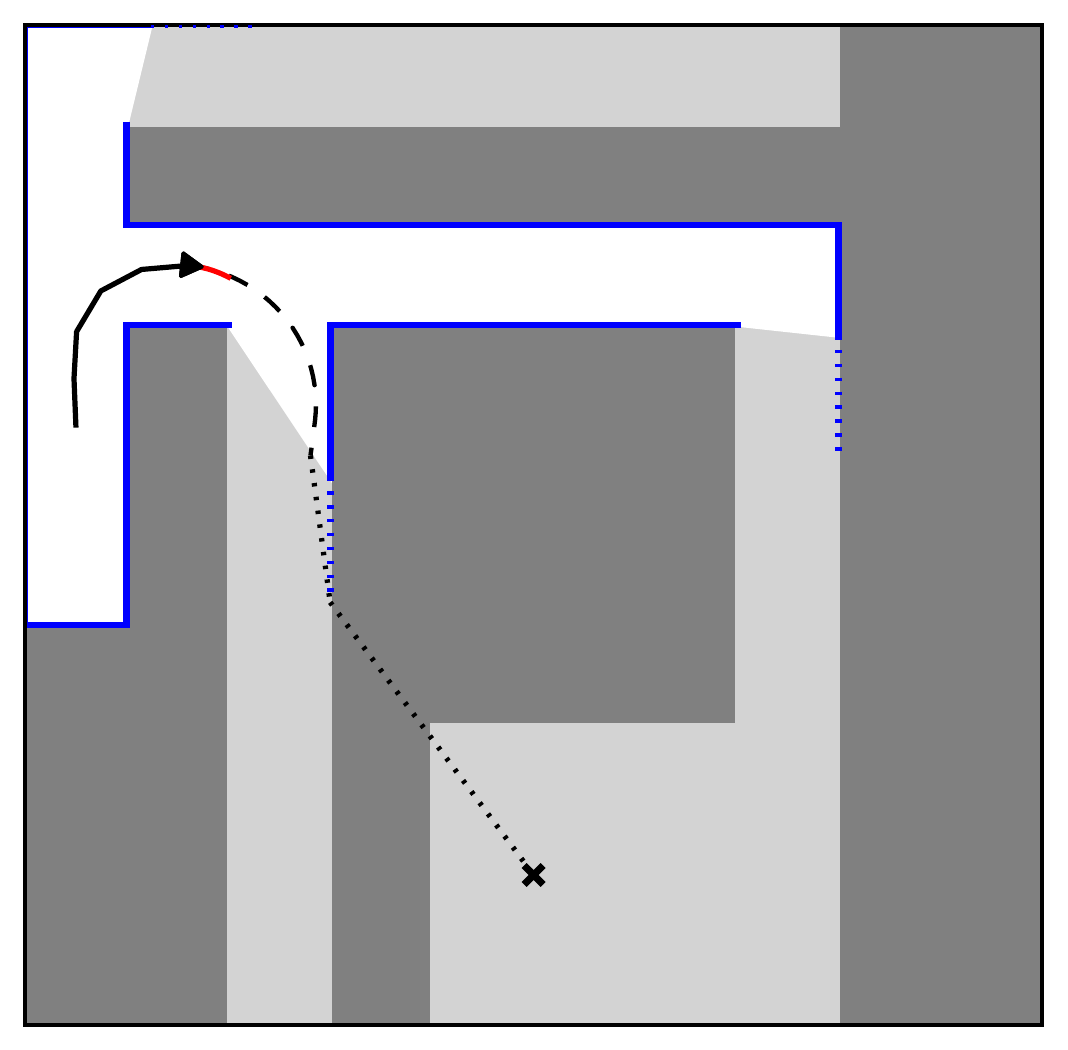}}
\subfigure[]{
\centering
\label{fig:maze_c}\includegraphics[width=0.3\linewidth]{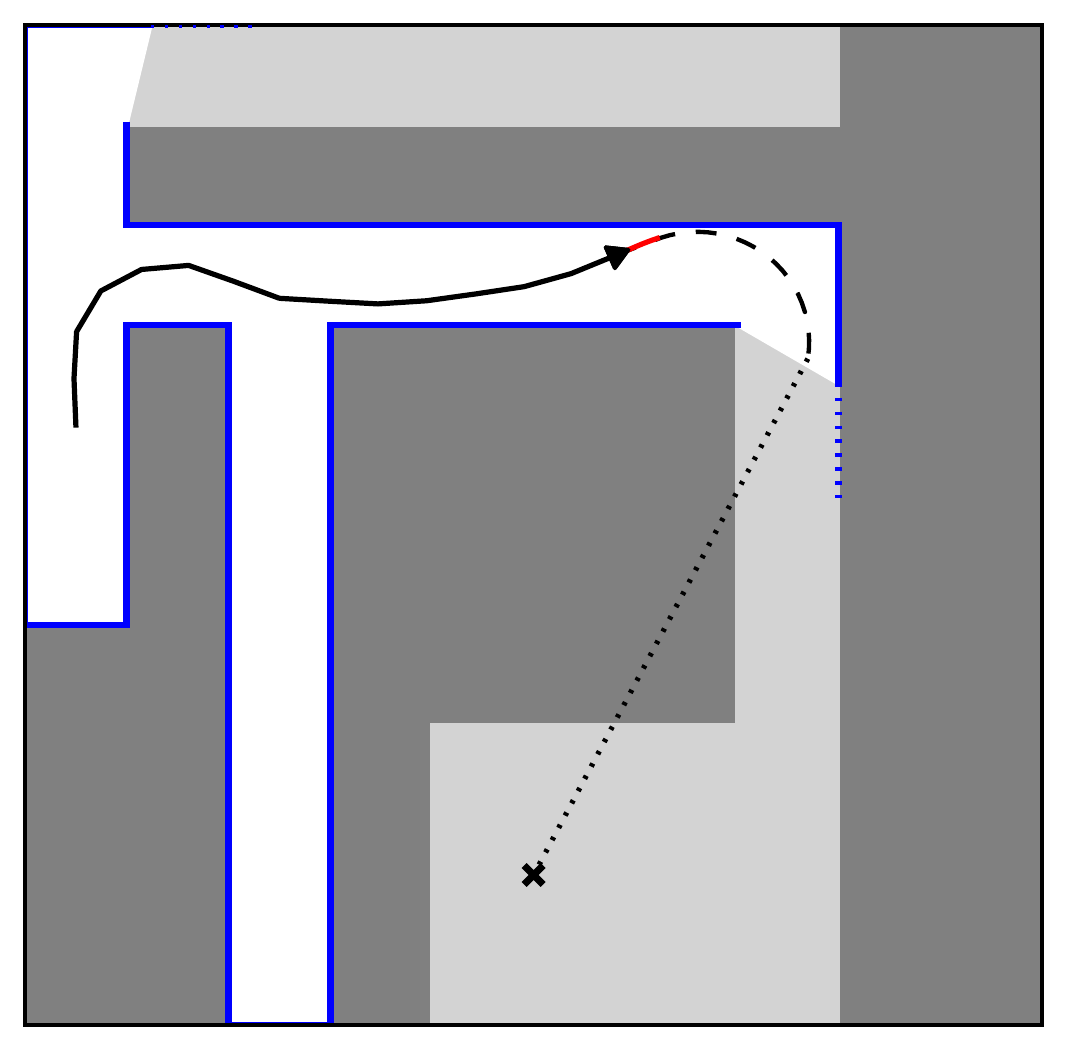}}
\subfigure[]{
\centering
\label{fig:maze_d}\includegraphics[width=0.3\linewidth]{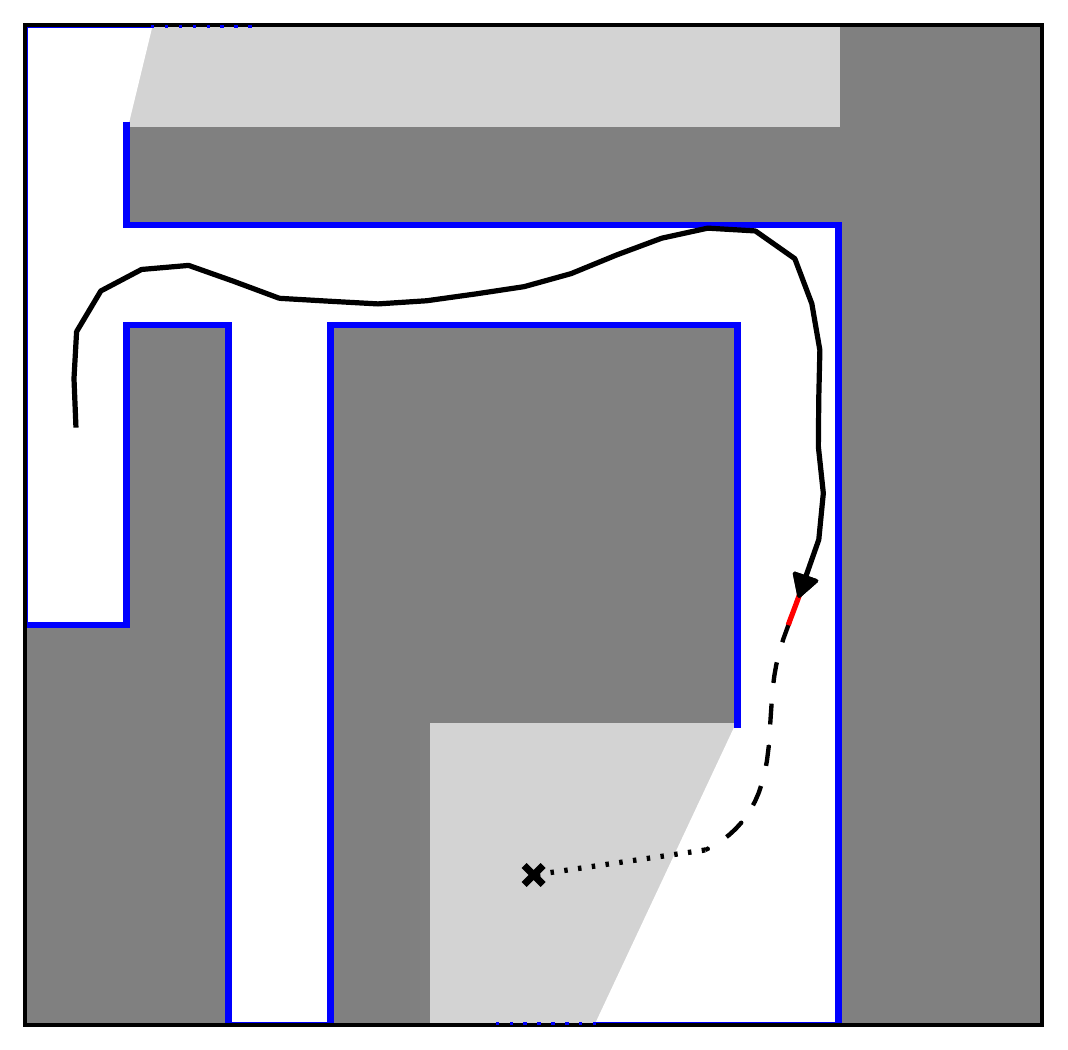}}
\subfigure[]{
\centering
\label{fig:maze_e}\includegraphics[width=0.75\linewidth]{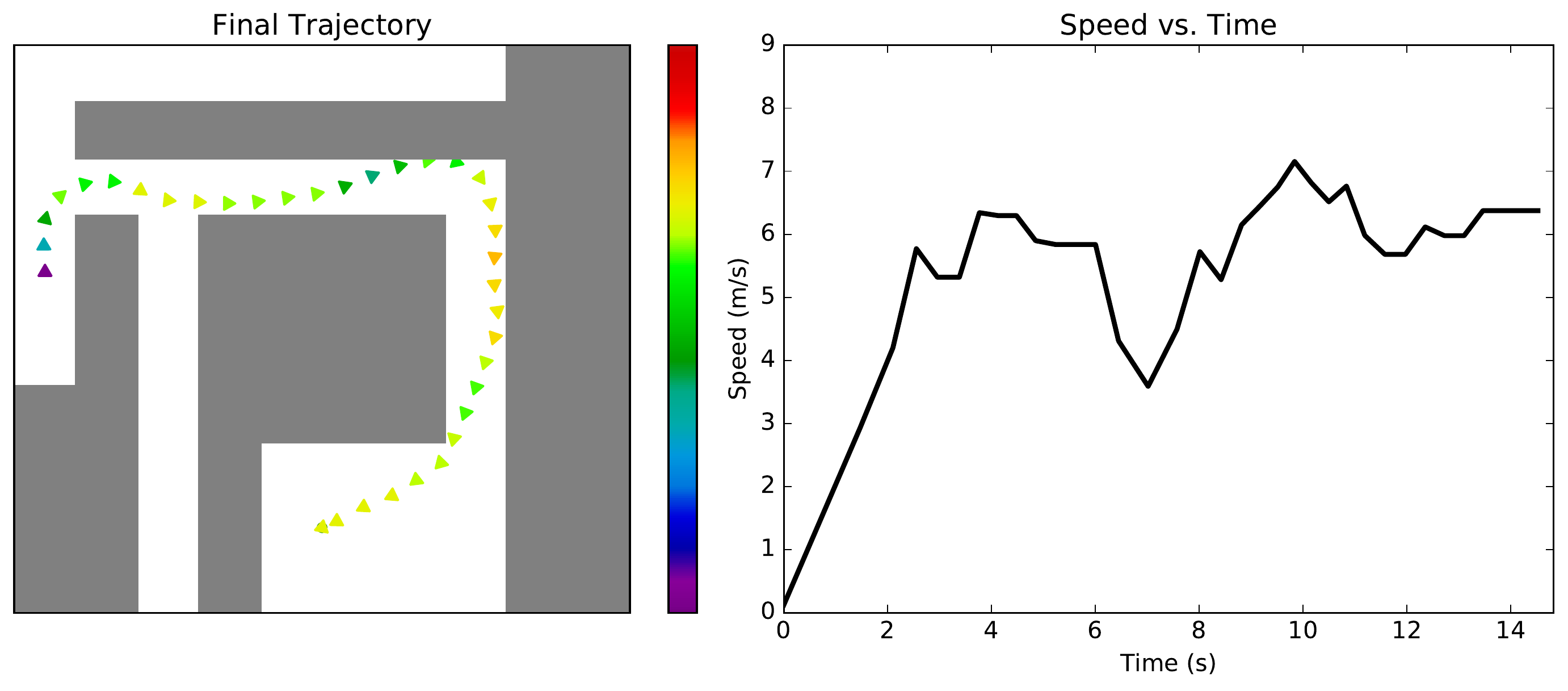}}
\caption{Simulation results on a maze map with dead ends and forks in the path. The receding horizon plan to the goal is shown in the dashed (current to intermediate goal) and the dotted (intermediate goal to final goal) lines. The travelled path up to each waypoint is shown in black, with the action to be taken at the timestep in red. The full path is shown in (d), with speed indicated by the color bar. Note the recovery from exploration of a hallway that turned out to be a dead end shown in (a) and (b). }
\label{fig:maze}
\end{figure}

\begin{figure}[t!]
\centering     
\subfigure[]{
\centering
\label{fig:forest_a}\includegraphics[width=0.3\linewidth]{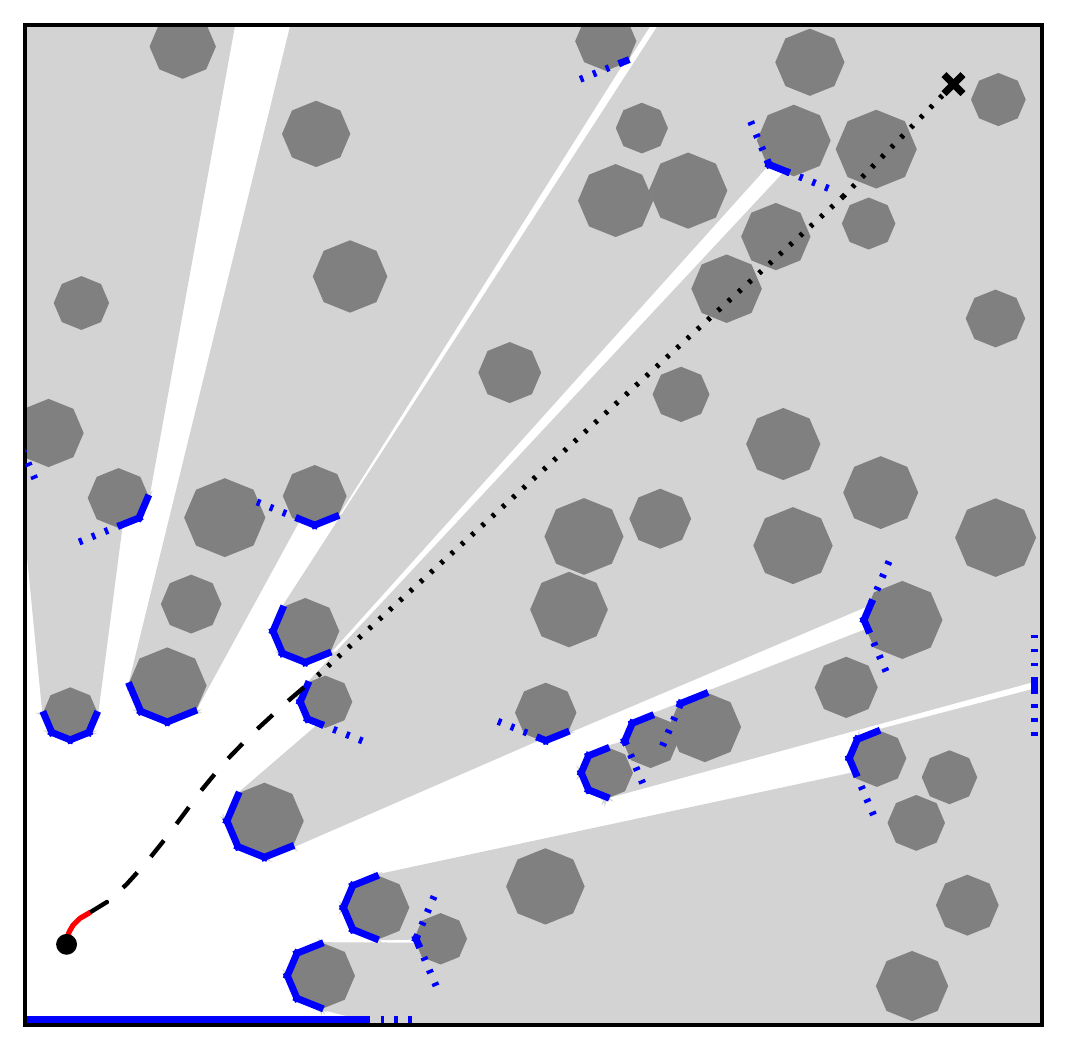}} 
\subfigure[]{
\centering
\label{fig:forest_b}\includegraphics[width=0.3\linewidth]{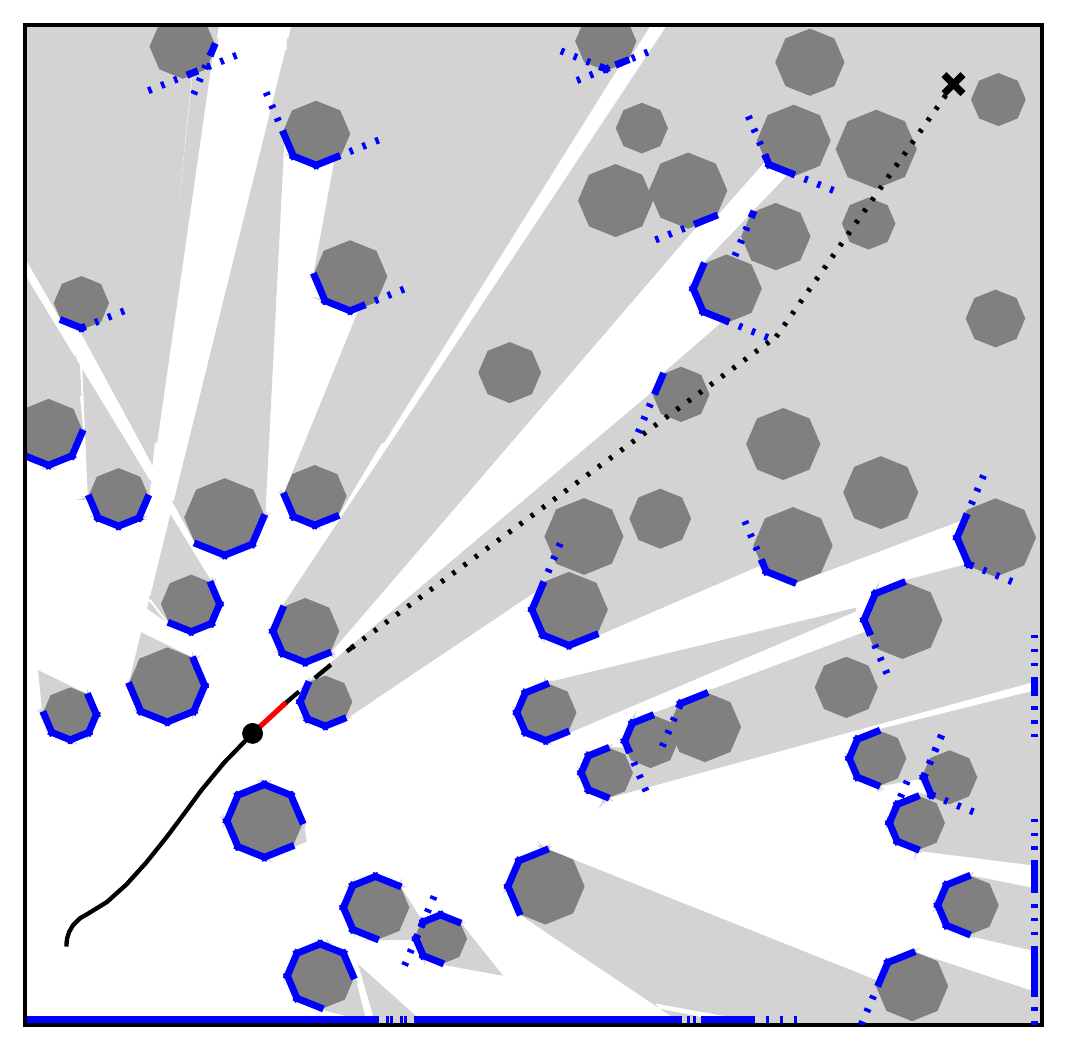}}
\subfigure[]{
\centering
\label{fig:forest_d}\includegraphics[width=0.3\linewidth]{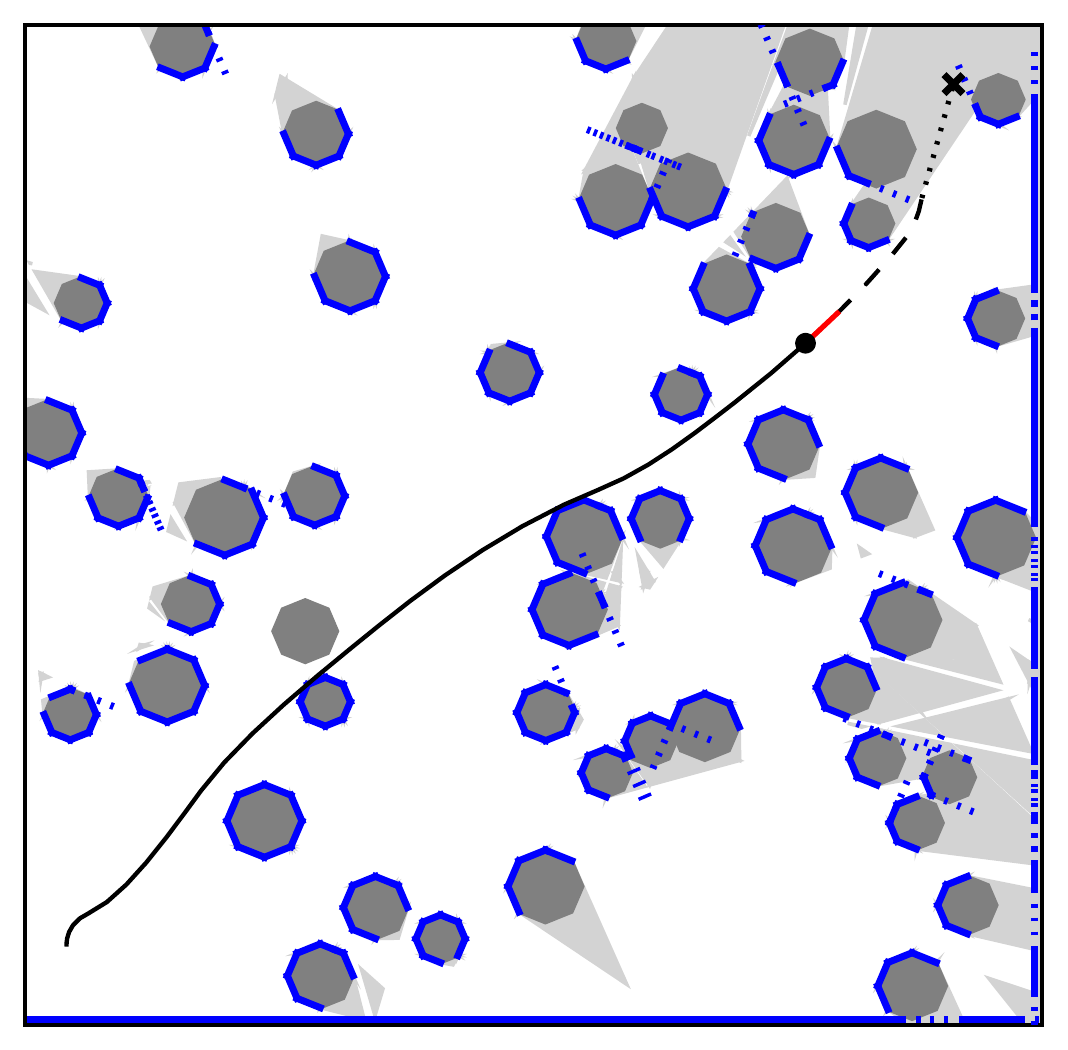}}
\subfigure[]{
\centering
\label{fig:forest_e}\includegraphics[width=0.75\linewidth]{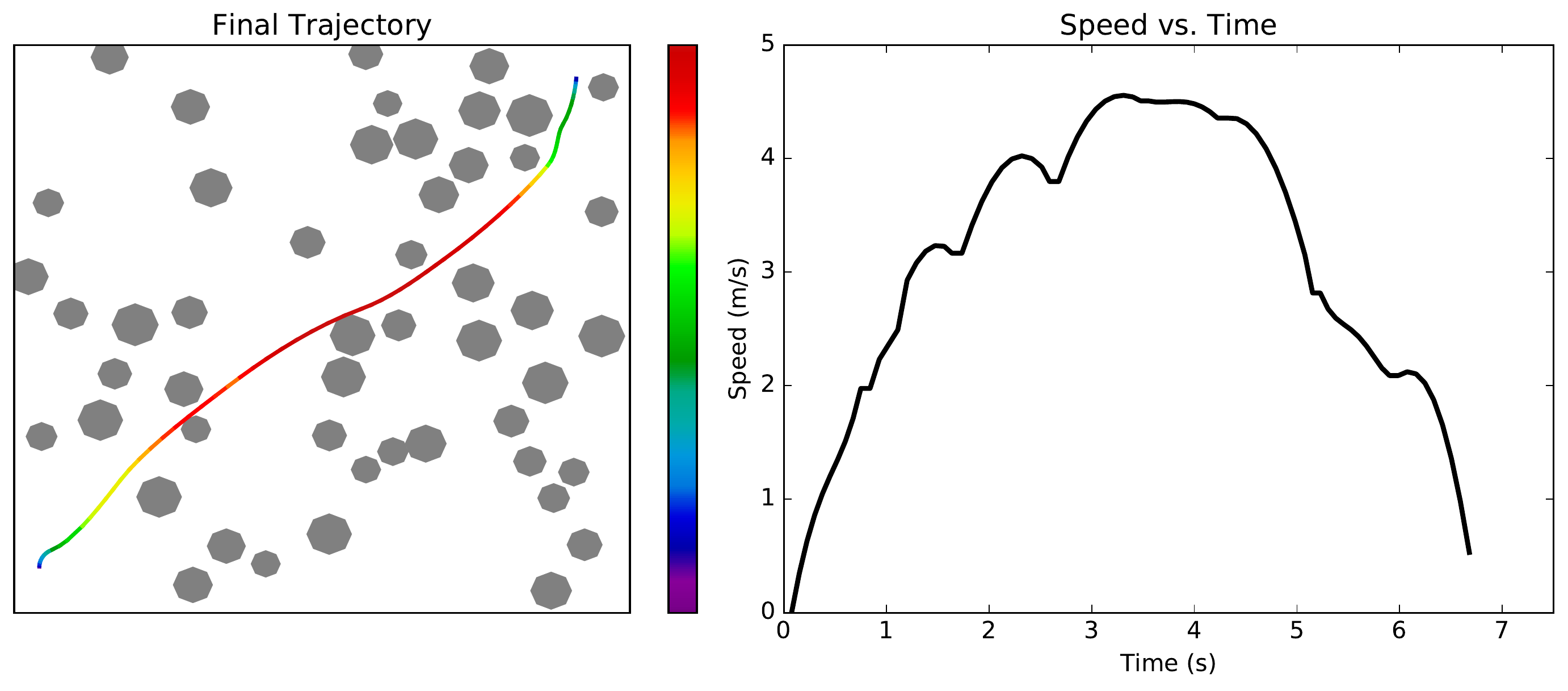}}
\caption{Simulation results on a forest environment with tree obstacles. Formatting is the same as in Figure \ref{fig:maze}.}
\label{fig:forest}
\end{figure}

\begin{figure}[t!]
\centering     
\subfigure[]{
\centering
\label{fig:gates_a}\includegraphics[width=0.3\linewidth]{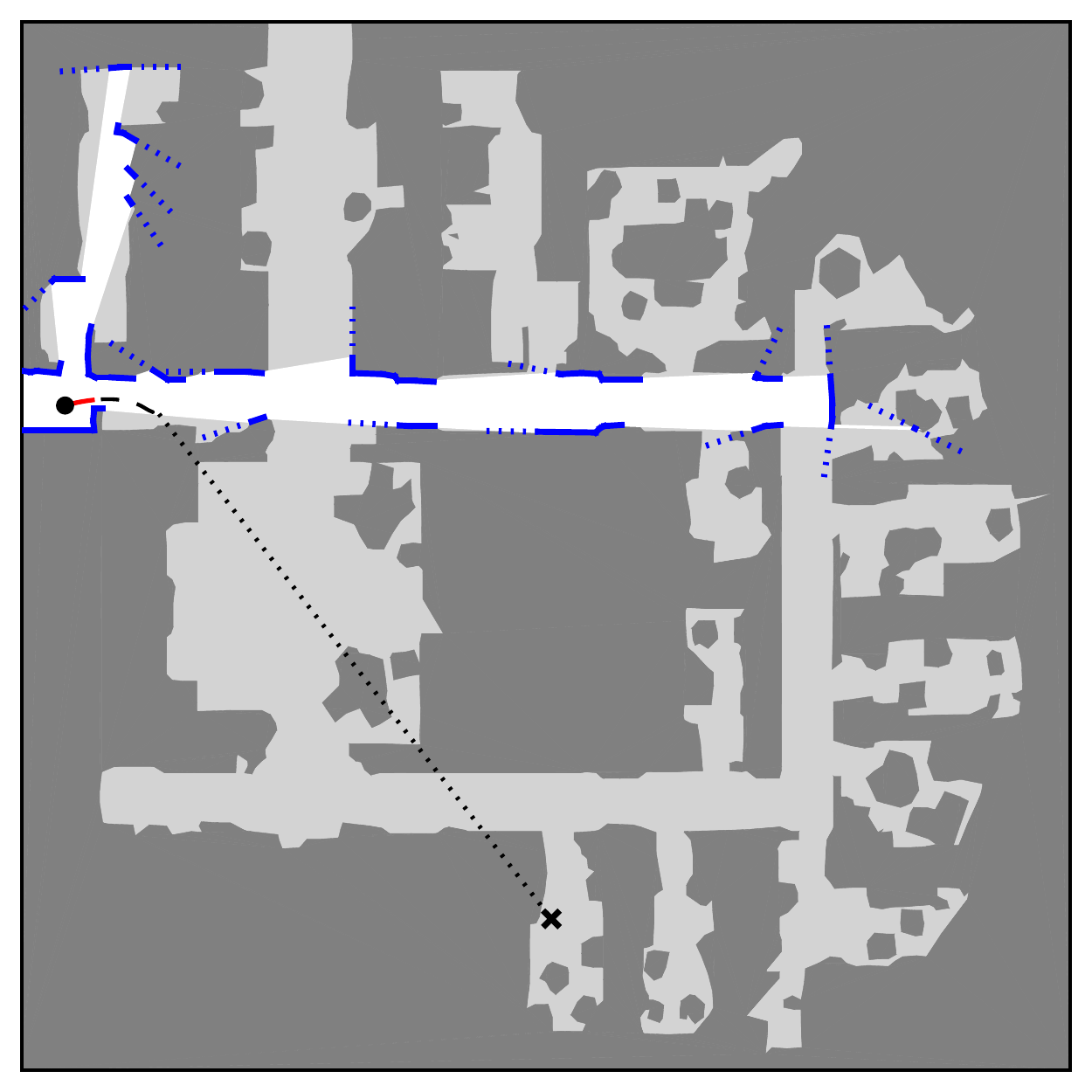}} 
\subfigure[]{
\centering
\label{fig:gates_c}\includegraphics[width=0.3\linewidth]{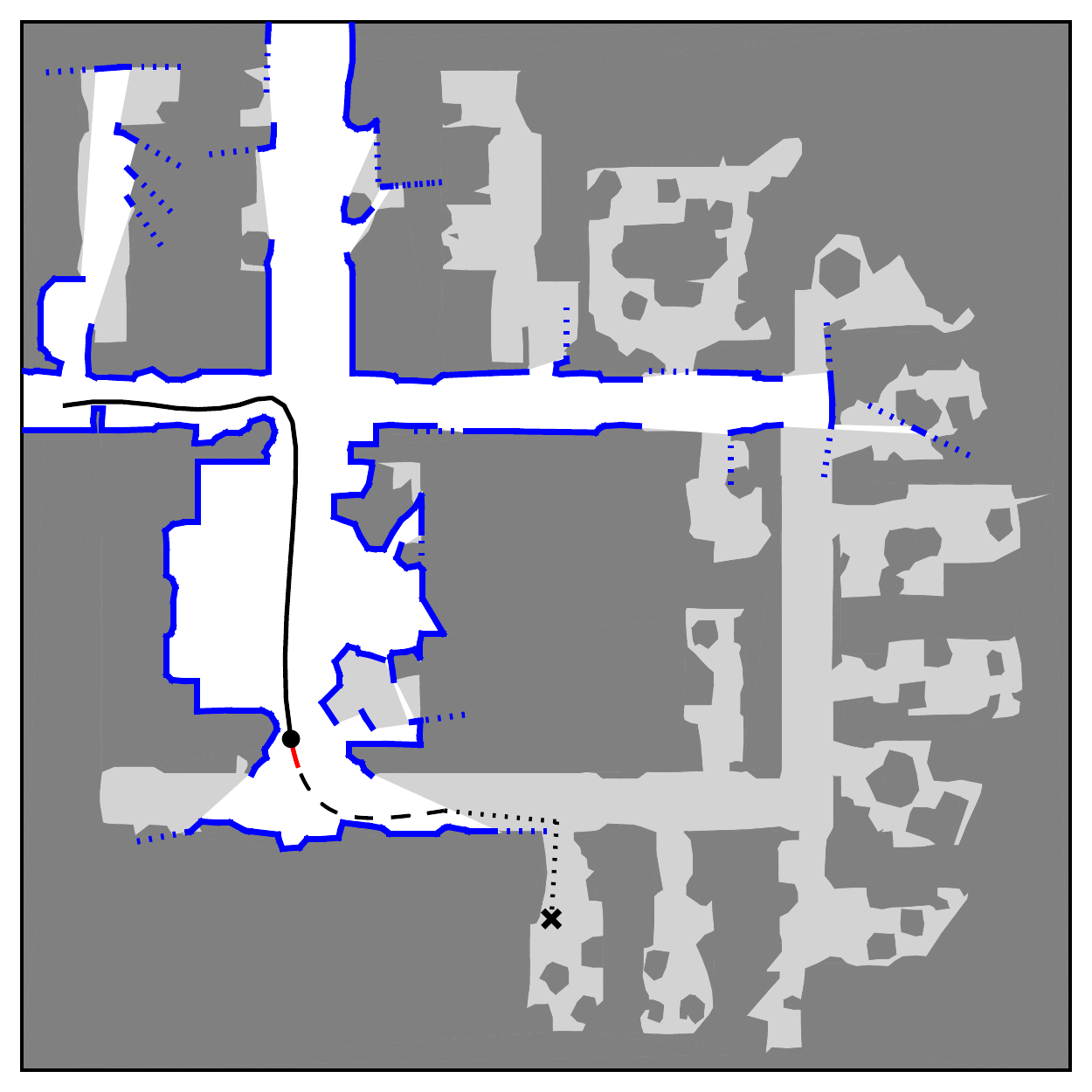}}
\subfigure[]{
\centering
\label{fig:gates_d}\includegraphics[width=0.3\linewidth]{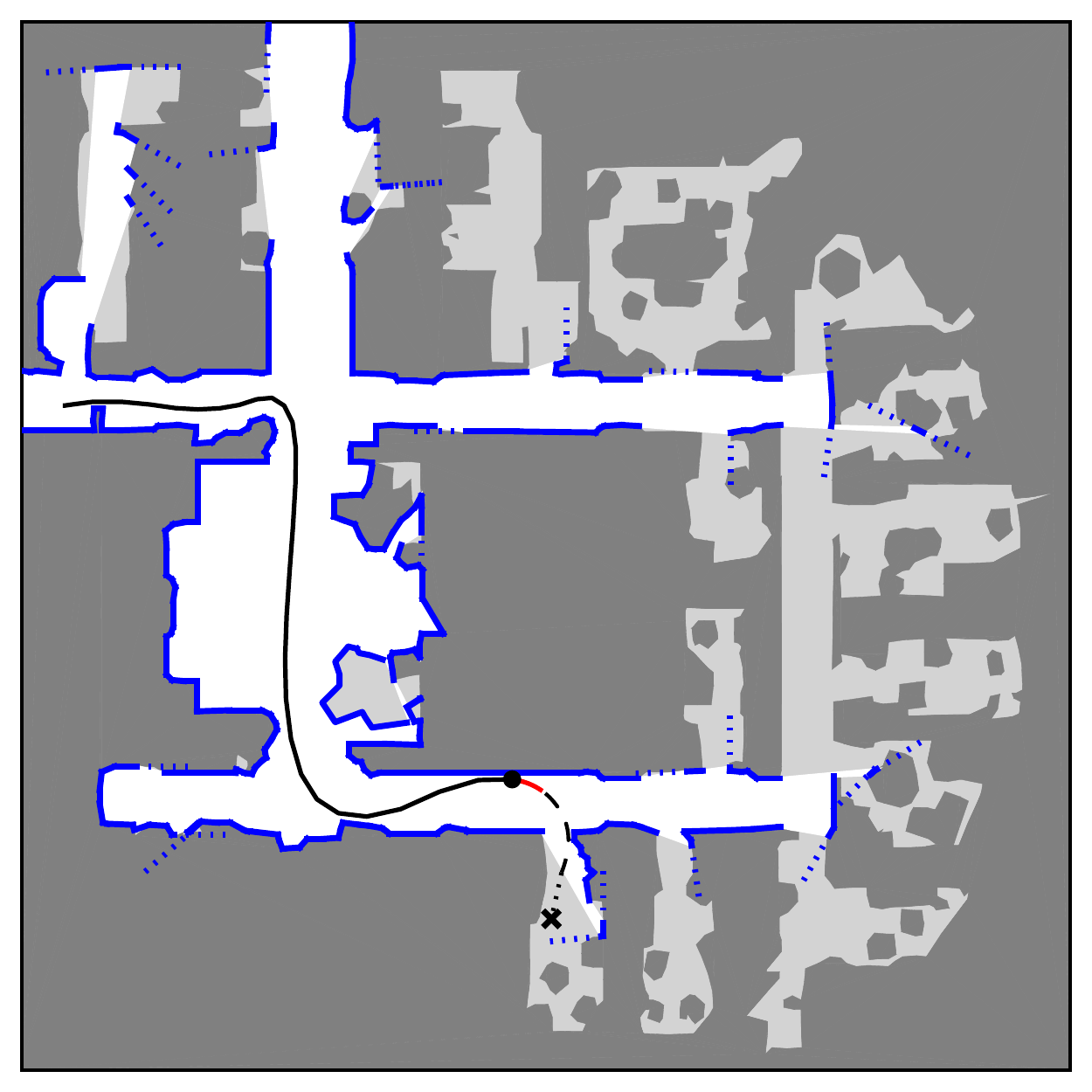}}
\subfigure[]{
\centering
\label{fig:gates_e}\includegraphics[width=0.75\linewidth]{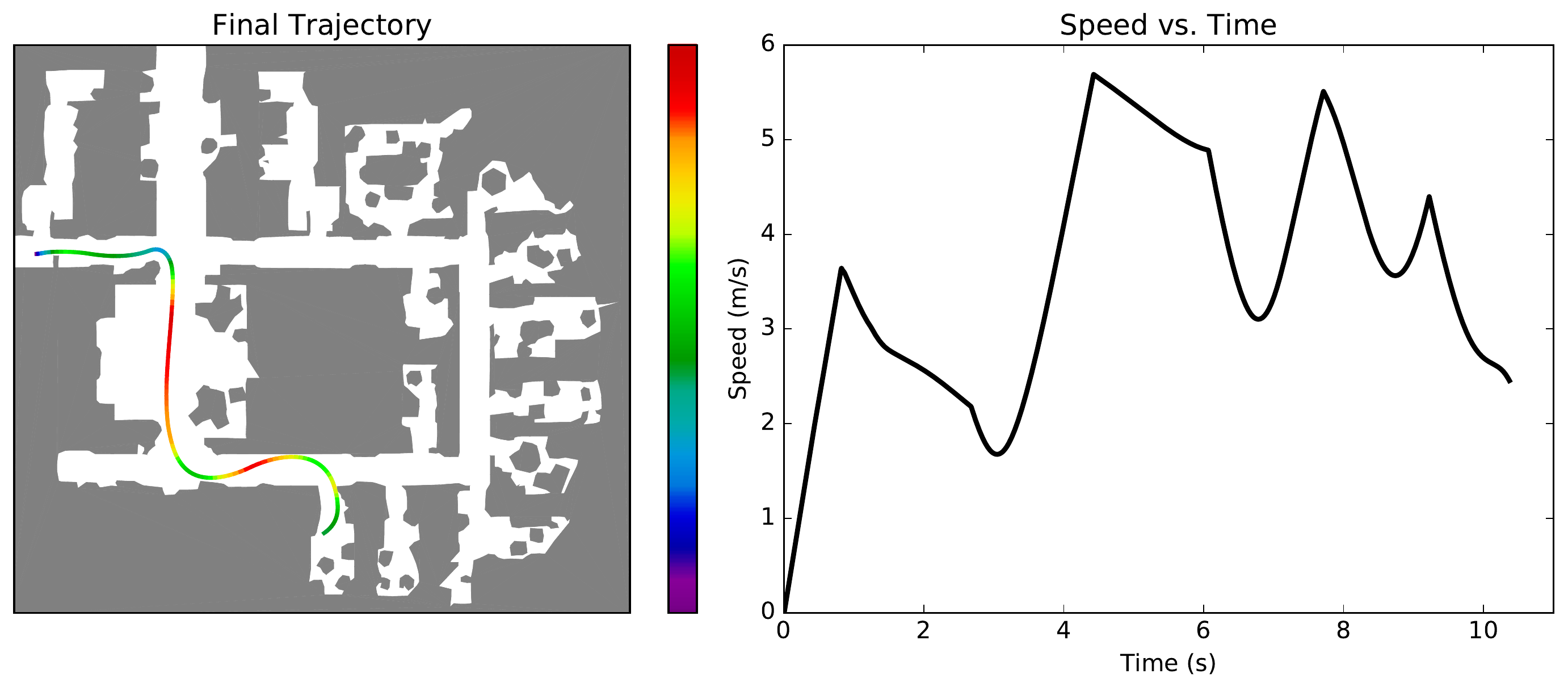}}
\caption{Simulation results on simplified scan of the Stanford Gates building internals. Formatting is the same as in Figure \ref{fig:maze}.}
\label{fig:gates}
\end{figure}

To understand better the behavior of our implementation of $\pola$, Figure \ref{fig:maze} displays the results of our algorithm on a maze environment with the nonlinear vehicle dynamics. We see that the vehicle takes wide turns when approaching corners, which allows for greater visibility of what is around the corner before actually making the turn and minimizes the deceleration necessary to maintain safety. The optimization {\em automatically and implicitly} trades off the extra distance required to take the turn more widely and the extra speed allowed by seeing the open hallway earlier and thus not having to slow down to ensure safety in case of a hidden obstacle. Figures~\ref{fig:maze_b} and \ref{fig:maze_c} show the added benefit of this approach---early detection of bad trajectories. While a myopic approach may have optimistically rushed into the path suggested in Figure~\ref{fig:maze_b}, by taking the turn more widely (which, again, was never explicitly programmed into our algorithm) the vehicle detects the dead end early and swings back up to the correct route, having suffered only a minor delay. 
Figures \ref{fig:forest} and \ref{fig:gates} show results of our algorithm with double integrator dynamics in a forest environment with scattered tree obstacles and a real office building environment with complex hallways, furniture, and rooms. In the forest environment, greedy exploration and plans to intermediate goals guide the robot to the goal on a direct and fast trajectory with minimal detours despite the lack of any prior knowledge of the environment (the $\upd$ function used was the same as for the maze) and guaranteed safety. The office building simulation shows each of the safety and optimality features of our algorithm. Initial travel in the narrow hallway is slow and cautious, followed by higher speeds in the wider space, and finally wide turns when traveling back into the narrower hallway and into the room to arrive at the goal.

Table~\ref{table:comp_times} shows the computation times for each environment with their corresponding complexity defined by number of half-planes defining the obstacles in the map and the average number of half-planes defining the unexplored frontier at the beginning of each action. Each action requires about 0.5 - 1s of (serial) computation time, which makes the algorithm amenable to real-time implementation. Computation times could be significantly improved by considering a parallel implementation, which is left for future work.

\newcolumntype{C}[1]{>{\centering\let\newline\\\arraybackslash\hspace{0pt}}m{#1}}
\newcolumntype{B}[1]{>{\centering\let\newline\\\arraybackslash\hspace{0pt}}b{#1}}
\begin{table}[t]
\centering
\caption{Computation Times}
\label{table:comp_times}
\begin{tabular}{l|C{2.2cm}|C{2.1cm}|C{1.3cm}}
 Environment & Total \# of obstacle half-planes & Avg. \# of frontier half-planes  & Avg. Time (ms)/plan \\ \hline 
 Maze & 96 & 2  & 540 \\
 Forest & 490 & 15  & 780  \\
 Office Bldg. & 1493 & 25  & 1340
\end{tabular}
\end{table}

\section{Conclusions and Future Work}\label{sec:conc}
We have proposed a algorithmic framework and a pseudo-optimal class of policies for the problem of safe motion planning in unknown environments. This work easily raises as many interesting questions as it solves, and future work will include (1) improving computation times through parallelization and optimized implementation, (2) deployment on real robot and integration with true dynamics and perception functions, (3) theoretical bounds on the optimality gap of $\polu$ as a function of $\upd$, (4) extension to the setting of uncertainty, both in the dynamics and perception of the robot, and (5) investigation into the best implementation details for different types of problems, especially the crucial choices of $\icsa$ and $\upd$. We highlight this last direction, the choice of $\upd$, as particularly appealing for its allowance of sophisticated machine learning algorithms to guess at the unexplored environment. While especially the most complex learning algorithms can have spectacular performance, they are often criticized for a lack of reliability and generalizability---but $\polu\,$/$\pola$ completely ameliorate this concern by acting as a wrapper around the machine learning in $\upd$ that guarantees the robot's safety no matter what it returns. Underscoring this point: if $\upd$ ever returns a grossly inaccurate guess for the environment due to one of the many reasons why machine learning algorithms can fail (e.g., test data which does not resemble the training data), the robot will (safely) begin to follow a less-efficient trajectory. This seems like as good an outcome as possible, since of course no algorithm can perform equally well if given good or bad information, but using the methods in this paper the worst we can lose is time, but never the robot itself due to collision.

\bibliographystyle{plainnat}
\bibliography{../../../bib/main,../../../bib/ASL_papers}

\newcommand{\noopsort}[1]{} \newcommand{\printfirst}[2]{#1}
  \newcommand{\singleletter}[1]{#1} \newcommand{\switchargs}[2]{#2#1}
\begin{thebibliography}{26}
\providecommand{\natexlab}[1]{#1}
\providecommand{\url}[1]{\texttt{#1}}
\expandafter\ifx\csname urlstyle\endcsname\relax
  \providecommand{\doi}[1]{doi: #1}\else
  \providecommand{\doi}{doi: \begingroup \urlstyle{rm}\Url}\fi

\bibitem[B.(1997)]{Yamauchi1997}
Yamauchi B.
\newblock A frontier-based approach for autonomous exploration.
\newblock In \emph{{Proc.\ IEEE Int.\ Symp.\ on Computational Intelligence in
  Robotics and Automation}}, 1997.

\bibitem[Bekris and Kavraki(2007)]{BekrisKavraki2007}
K.~E. Bekris and L.~E. Kavraki.
\newblock Greedy but safe replanning under kinodynamic constraints.
\newblock In \emph{{IEEE Conf.\ on Robotics and Automation}}, 2007.

\bibitem[Bezanson et~al.(2012)Bezanson, Karpinski, Shah, and
  Edelman]{BezansonKarpinskiEtAl2012}
J.~Bezanson, S.~Karpinski, V.~B. Shah, and A.~Edelman.
\newblock Julia: A fast dynamic language for technical computing, 2012.
\newblock {Available} at \url{http://arxiv.org/abs/1209.5145}.

\bibitem[Bircher et~al.(2016)Bircher, Kamel, Alexis, Oleynikova, and
  Siegwart]{BircherKamelEtAl2016}
A.~Bircher, M.~Kamel, K.~Alexis, H.~Oleynikova, and R.~Siegwart.
\newblock Receding horizon {"Next-Best-View"} planner for {3D} exploration.
\newblock In \emph{{IEEE Conf.\ on Robotics and Automation}}, 2016.

\bibitem[Borodin and El-Yaniv(2005)]{BorodinEl-Yaniv05}
A.~Borodin and R.~El-Yaniv.
\newblock \emph{Online computation and competitive analysis}.
\newblock {Cambridge University Press}, 2005.

\bibitem[Burgard et~al.(2005)Burgard, Moors, Stachniss, and
  Schneider]{BurgardMoorsEtAl2005}
W.~Burgard, M.~Moors, C.~Stachniss, and F.~E. Schneider.
\newblock Coordinated multi-robot exploration.
\newblock \emph{{IEEE Transactions on Robotics}}, 21\penalty0 (3):\penalty0
  376--386, 2005.

\bibitem[Connolly(1985)]{Connolly1985}
C.~Connolly.
\newblock The determination of next best views.
\newblock In \emph{{Proc.\ IEEE Conf.\ on Robotics and Automation}}, volume~2,
  pages 432--435, 1985.

\bibitem[Ferguson et~al.(2006)Ferguson, Kalra, and
  Stentz]{FergusonKalraEtAl2006}
D.~Ferguson, N.~Kalra, and A.~Stentz.
\newblock Replanning with {RRTs}.
\newblock In \emph{{Proc.\ IEEE Conf.\ on Robotics and Automation}}, 2006.

\bibitem[Fleischer et~al.(2008)Fleischer, Kamphans, Klein, Langetepe, and
  Trippen]{FleischerKamphansEtAl2008}
R.~Fleischer, T.~Kamphans, R.~Klein, E.~Langetepe, and G.~Trippen.
\newblock Competitive online approximation of the optimal search ratio.
\newblock \emph{{SIAM Journal on Computing}}, 38\penalty0 (3):\penalty0
  881--898, 2008.

\bibitem[Fraichard and Asama(2004)]{FraichardAsama2004}
T.~Fraichard and H.~Asama.
\newblock Inevitable collision states -- a step towards safer robots?
\newblock \emph{{Advanced Robotics}}, 18\penalty0 (10):\penalty0 1001--1024,
  2004.

\bibitem[Frazzoli et~al.(2002)Frazzoli, Dahleh, and
  Feron]{FrazzoliDahlehEtAl2002}
E.~Frazzoli, M.~A. Dahleh, and E.~Feron.
\newblock Real-time motion planning for agile autonomous vehicles.
\newblock \emph{{AIAA Journal of Guidance, Control, and Dynamics}}, 25\penalty0
  (1):\penalty0 116--129, 2002.

\bibitem[Heng et~al.(2015)Heng, Gotovos, Krause, and
  Pollefeys]{HengGotovosEtAl2015}
L.~Heng, A.~Gotovos, A.~Krause, and M.~Pollefeys.
\newblock Efficient visual exploration and coverage with a micro aerial vehicle
  in unknown environments.
\newblock In \emph{{Proc.\ IEEE Conf.\ on Robotics and Automation}}, pages
  1071--1078, 2015.

\bibitem[Hsu et~al.(2002)Hsu, Kindel, Latombe, and Rock]{HsuKindelEtAl2002}
D.~Hsu, R.~Kindel, J.-C. Latombe, and S.~Rock.
\newblock Randomized kinodynamic motion planning with moving obstacles.
\newblock \emph{{Int.\ Journal of Robotics Research}}, 21\penalty0
  (3):\penalty0 233--255, 2002.

\bibitem[Janson et~al.(2015)Janson, Schmerling, Clark, and
  Pavone]{JansonSchmerlingEtAl2015}
L.~Janson, E.~Schmerling, A.~Clark, and M.~Pavone.
\newblock {Fast} {Marching} {Tree:} a fast marching sampling-based method for
  optimal motion planning in many dimensions.
\newblock \emph{{Int.\ Journal of Robotics Research}}, 34\penalty0
  (7):\penalty0 883--921, 2015.

\bibitem[Karaman and Frazzoli(2011)]{KaramanFrazzoli2011}
S.~Karaman and E.~Frazzoli.
\newblock Sampling-based algorithms for optimal motion planning.
\newblock \emph{{Int.\ Journal of Robotics Research}}, 30\penalty0
  (7):\penalty0 846--894, 2011.

\bibitem[Koutsoupias and Papadimitriou(2000)]{KoutsoupiasPapadimitriou00}
E.~Koutsoupias and C.~H. Papadimitriou.
\newblock Beyond competitive analysis.
\newblock \emph{{SIAM Journal on Computing}}, 30\penalty0 (1):\penalty0
  300--317, 2000.

\bibitem[LaValle(2006)]{LaValle2006}
S.~M. LaValle.
\newblock \emph{Planning Algorithms}.
\newblock {Cambridge University Press}, 2006.

\bibitem[LaValle(2011)]{LaValle2011}
S.~M. LaValle.
\newblock Motion planning: Wild frontiers.
\newblock \emph{{IEEE Robotics and Automation Magazine}}, 18\penalty0
  (2):\penalty0 108--118, 2011.

\bibitem[LaValle and Kuffner(2001)]{LaValleKuffner2001}
S.~M. LaValle and J.~J. Kuffner.
\newblock Randomized kinodynamic planning.
\newblock \emph{{Int.\ Journal of Robotics Research}}, 20\penalty0
  (5):\penalty0 378--400, 2001.

\bibitem[Li et~al.(2016)Li, Littlefield, and Bekris]{LiLittlefieldEtAl2016}
Y.~Li, Z.~Littlefield, and K.~E. Bekris.
\newblock Asymptotically optimal sampling-based kinodynamic planning.
\newblock \emph{{Int.\ Journal of Robotics Research}}, 35\penalty0
  (5):\penalty0 528–564, 2016.

\bibitem[Petti and Fraichard(2005)]{PettiFraichard2005}
S.~Petti and T.~Fraichard.
\newblock Partial motion planning framework for reactive planning within
  dynamic environments.
\newblock In \emph{Proc.\ of the IFAC/AAAI Int.\ Conf.\ on Informatics in
  Control, Automation and Robotics}, 2005.

\bibitem[Richter et~al.(2018)Richter, Vega-Brown, and Roy]{RichterVega2018}
C.~Richter, W.~Vega-Brown, and N.~Roy.
\newblock Bayesian learning for safe high-speed navigation in unknown
  environments.
\newblock In \emph{{Int.\ Journal of Robotics Research}}, pages 325--341. 2018.

\bibitem[Schmerling et~al.(2015{\natexlab{a}})Schmerling, Janson, and
  Pavone]{SchmerlingJansonEtAl2015}
E.~Schmerling, L.~Janson, and M.~Pavone.
\newblock Optimal sampling-based motion planning under differential
  constraints: the driftless case.
\newblock In \emph{{Proc.\ IEEE Conf.\ on Robotics and Automation}},
  2015{\natexlab{a}}.
\newblock Extended version available at \url{http://arxiv.org/abs/1403.2483/}.

\bibitem[Schmerling et~al.(2015{\natexlab{b}})Schmerling, Janson, and
  Pavone]{SchmerlingJansonEtAl2015b}
E.~Schmerling, L.~Janson, and M.~Pavone.
\newblock Optimal sampling-based motion planning under differential
  constraints: the drift case with linear affine dynamics.
\newblock In \emph{{Proc.\ IEEE Conf.\ on Decision and Control}},
  2015{\natexlab{b}}.

\bibitem[van~den Berg and Overmars(2006)]{BergOvermars06}
J.~P. van~den Berg and M.~H. Overmars.
\newblock Planning the shortest safe path amidst unpredictably moving
  obstacles.
\newblock In \emph{{Workshop on Algorithmic Foundations of Robotics}}, pages
  103--118, 2006.

\bibitem[Webb and van~den Berg(2013)]{WebbBerg2013}
D.~J. Webb and J.~van~den Berg.
\newblock {Kinodynamic RRT*}: Optimal motion planning for systems with linear
  differential constraints.
\newblock In \emph{{Proc.\ IEEE Conf.\ on Robotics and Automation}}, 2013.

\end{thebibliography}

\end{document}